\ifdefined\XeTeXversion\else\ifdefined\pdfoutput\pdfoutput=1\fi\fi
\documentclass{article}

\usepackage[T1]{fontenc}
\usepackage{microtype}
\usepackage{graphicx}
\usepackage{subcaption}
\usepackage{booktabs}
\usepackage{xurl}
\usepackage{hyperref}

\usepackage[preprint]{icml2027}
\makeatletter
\renewcommand{\ICML@preprint}{\textit{Preprint. November 20, 2026.}}
\makeatother
\hypersetup{pdfsubject={Preprint}}

\usepackage{amsmath}
\usepackage{amssymb}
\usepackage{mathtools}
\usepackage{amsthm}
\usepackage{multirow}
\usepackage{placeins}
\usepackage{makecell}
\usepackage[capitalize,noabbrev]{cleveref}
\usepackage{algorithm}
\usepackage{algorithmic}

\newcommand{\Cphys}{C_{\mathrm{phys}}}
\newcommand{\Cctrl}{C_{\mathrm{ctrl}}}
\newcommand{\Cvio}{C_{\mathrm{VIO}}}
\newcommand{\ploc}{p_{\mathrm{loc}}}
\newcommand{\pvio}{p_{\mathrm{VIO}}}
\newcommand{\pfb}{p_{\mathrm{fb}}}
\newcommand{\pmap}{p_{\mathrm{map}}}
\newcommand{\pref}{p_{\mathrm{ref}}}
\newcommand{\rrec}{r_{\mathrm{rec}}}
\newcommand{\Tb}{\textsc{T8}}
\newcommand{\Sl}{\textsc{Slong}}
\newcommand{\Ss}{\textsc{Sshort}}
\newcommand{\Tf}{\textsc{T4}}
\newcommand{\HT}{\mathcal{H}_T}
\newcommand{\HS}{\mathcal{H}_S}
\newcommand{\Zzone}{\mathcal{Z}}
\newcommand{\pp}{\,pp}
\newcommand{\1}{\mathbb{I}}

\theoremstyle{plain}
\newtheorem{lemma}{Lemma}
\newtheorem{remark}{Remark}

\icmltitlerunning{Portable Active Recovery under Visual Outages (Preprint)}

\begin{document}

\twocolumn[
  \icmltitle{Portable Active Recovery under Timed and\texorpdfstring{\\}{ }Position-Dependent Visual Outages}

   \begin{icmlauthorlist}
    \icmlauthor{Jinyuan Zhang}{affil1}
    \icmlauthor{Shengshuo Jiao}{affil2}
  \end{icmlauthorlist}
  \begin{center}
    \vskip -0.05in
    {\small\texttt{\{jinyuanzhang, 202621120012764\}@stu.hubu.edu.cn}}
  \end{center}
  \icmlaffiliation{affil1}{School of Computer Science, Hubei University, Wuhan 430062, China;}
  \icmlaffiliation{affil2}{School of Cyber Science and Technology, Hubei University, Wuhan 430062, China}
  \icmlcorrespondingauthor{Jinyuan Zhang}{jinyuanzhang@stu.hubu.edu.cn}

  \icmlkeywords{visual-inertial odometry, active perception, decision making under partial observability, quadrotor control, runtime assurance}

  \vskip 0.3in
]

\printAffiliationsAndNotice{}

\begin{abstract}
When a camera blackout depends on physical position rather than elapsed time, stable hovering prevents visual recovery and mission completion. On stereo visual-inertial quadrotors without external localization, an exogenous \emph{timed} blackout and an endogenous \emph{position-dependent} dark zone appear identical at loss onset yet require opposite responses: hover-and-wait completes 93/96 timed blackouts but 0/96 spatial outages (13 hard failures), whereas blind continuation traverses many spatial zones yet fails heavily during timed blackouts. We formulate visual-outage recovery as a worst-condition active-observability problem and introduce a portable recovery module at the reference interface of an unchanged host controller with zero retraining. Using deployable telemetry alone, the module coordinates visual-loss detection, jerk-limited braking, bounded backward probing, settled rejoining, height-integral trim, and an optional gated prior-map branch. On a preregistered 48-scene benchmark (1{,}920 flights), our full MPPI configuration reaches 88.5\% [81.2, 94.8] worst-condition completion, outperforming development-selected DATT-L1 by $+29.2$\pp{} [17.7, 40.6] and exceeding all six external baselines after Bonferroni correction. Attached without retraining to SimpleFlight, the module lifts worst-condition completion by $+17.7$\pp{} [7.3, 28.1] to 86.5\% (and 90\% across $1.0\times/1.25\times$ speeds, 0/80 hard failures). Trace diagnostics reveal that survival hinges on visual-inertial re-acquisition asymmetry at vision return---not blind IMU drift suppression---and uncover a bias--drift map crossover confirmed on 72 held-out EuRoC windows.
\end{abstract}

\section{Introduction}
\label{sec:intro}

When a stereo visual-inertial quadrotor loses usable camera frames, \emph{stable hovering is not task recovery}. Modern learned and adaptive flight controllers track agile trajectories with high accuracy while pose estimates remain available~\citep{kaufmann2023champion,song2023racing,oconnell2022neuralfly,huang2023datt,chen2025simpleflight}, yet multi-second visual blackouts force open-loop inertial propagation while the tracker keeps acting on drifting states.

Under deployable onboard telemetry, two outage regimes are observationally aliased at loss onset yet impose opposite recovery laws. In an exogenous \emph{timed} blackout (\Tb{}, 8\,s), camera frames return automatically once the interval expires; stopping on inertial propagation preserves local visual context and completes 93/96 flights (\cref{sec:experiments}). In an endogenous \emph{position-dependent} dark zone (\Sl{}, 1.35\,m arc), frames return only after the vehicle physically leaves the unobservable region; hovering inside the zone completes 0/96 flights (13 hard failures). Conversely, continuing blindly along the schedule traverses many spatial zones yet incurs 7--12 timed-outage hard failures across external trackers. Completing missions reliably when the outage mechanism is undisclosed requires an active-observability decision layer~\citep{bajcsy1988active,kaelbling1998pomdp} above low-level tracking, analogous to Simplex runtime assurance~\citep{sha2001simplex}.

We decouple active recovery from low-level actuation at the reference interface, enabling zero-retraining deployment above unchanged host controllers. Beyond closed-loop recovery performance, trace-level diagnostics overturn a widely held intuition: stopping succeeds \emph{not} by slowing open-loop IMU drift---which grows at a similar quadratic rate across actions and hosts ($\le 1$\,cm difference at 4\,s despite $4.4\times$ higher body-rate excitation on MPPI)---but by governing \emph{visual-inertial re-acquisition} when stereo frames return. Stopping or probing back into previously observed texture enables Basalt's KLT optical flow to collapse accumulated blind error to 2--8\%, whereas forward traversal onto unseen far-side texture induces host-dependent re-acquisition jumps that a gated prior-map branch corrects on MPPI.

\textbf{Contributions.} This paper makes three contributions:
\begin{enumerate}
\setlength\itemsep{0.5pt}
\item \textbf{Worst-condition aliased-outage formulation and kinematic analysis (\cref{sec:problem}):} We formalize timed versus position-dependent visual-outage recovery under parallel physical ($\Cphys$) and localization-qualified ($\Cctrl$, $\Cvio$) task-completion criteria scored by worst-condition completion $W$, accompanied by a 1D kinematic lemma establishing exact quintic stopping depth, probe-budget feasibility, and two-segment blind-exposure reduction.
\item \textbf{Portable reference-interface recovery module (\cref{sec:method}):} Operating solely on deployable telemetry and references without altering host weights, our module coordinates visual-loss detection, jerk-limited braking, bounded backward probing (0.50\,m planned budget), settled rejoining, height-integral trim, and an optional support- and consistency-gated ORB-SLAM3 translation branch across three distinct flight controllers.
\item \textbf{Preregistered benchmark, cross-host transfer, and mechanistic diagnostics (\cref{sec:experiments}):} Across 48 fresh scenes (1{,}920 flights), the full MPPI configuration reaches 88.5\% [81.2, 94.8] worst-condition $\Cctrl$, beating development-selected DATT-L1 by $+29.2$\pp{} [17.7, 40.6] (H1) and all six external baselines under Bonferroni correction (99.17\%). Without retraining, the module lifts \Tb{} completion across MPPI, SimpleFlight, and DATT-L1, raises SimpleFlight worst-condition completion by $+17.7$\pp{} [7.3, 28.1] to 86.5\% (and 90\% across $1.0\times/1.25\times$ speeds with 0/80 hard failures), and uncovers the re-acquisition asymmetry and bias--drift map crossover confirmed on real EuRoC data.
\end{enumerate}

\section{Related Work}
\label{sec:related}

\textbf{Learned and adaptive quadrotor control.} Learned trackers---imitation~\citep{loquercio2021highspeed} and deep RL~\citep{kaufmann2023champion,song2023racing,chen2025simpleflight}---adaptive controllers~\citep{oconnell2022neuralfly,huang2023datt}, $\mathcal{L}_1$ augmentation~\citep{hovakimyan2010l1,wu2022l1quad}, and differentiable-simulation adaptation~\citep{pan2026lotf} excel at rejecting aerodynamic mismatch under continuous pose feedback. \citet{engelsman2026certainty} incorporate estimator uncertainty directly into nonlinear feedback laws to address transient estimation--control coupling, including sensor dropouts. Our approach complements low-level control design by operating purely at the reference interface of unchanged host controllers, enabling active recovery across four distinct controller families (MPPI~\citep{williams2017mppi}, SimpleFlight, DATT-L1/L1-MPPI, and LOTF) under multi-second visual blackouts.

\textbf{Visual-inertial estimation and map relocalization.} Visual-inertial odometry~\citep{mourikis2007msckf,forster2017preintegration,qin2018vinsmono,geneva2020openvins,usenko2020basalt} propagates the IMU open loop during blackouts, learned inertial odometry~\citep{liu2020tlio,cioffi2023imo,cioffi2023hdvio,qiu2025airio} lowers inertial drift, and prior-map relocalization~\citep{lynen2015getout,cadena2016slam,campos2021orbslam3} anchors pose once visual matches return. Existing studies primarily evaluate open-loop estimation accuracy; here we leave Basalt VIO unmodified, add a support- and consistency-gated ORB-SLAM3 translation branch, and quantify both closed-loop mission survival and open-loop real-data behavior on EuRoC~\citep{burri2016euroc}.

\textbf{Active observability, hidden-model POMDPs, and SLAM recovery.} Active perception~\citep{bajcsy1988active} and belief-space planning~\citep{kaelbling1998pomdp,platt2010belief} choose motion to acquire information; perception-aware MPC keeps textured regions in view during nominal flight~\citep{falanga2018pampc,zhang2018perception}, and monocular recovery re-initializes after throws~\citep{faessler2015recovery}. In SLAM failure recovery, \citet{qureshi2024onestep} train learned backtracking moves to restore tracking, while subterranean field teams coordinate multi-modal fallbacks under perceptual degradation~\citep{tranzatto2024cerberus}. Recent POMDP theory formalizes discrete model uncertainty~\citep{bovy2025mepomdp,galesloot2025hmpomdp} and active measurement selection~\citep{krale2023atm}. Timed and spatial visual blackouts share this hidden-model structure at loss onset: our bounded backward probe actively tests the outage regime while returning to seen texture to re-anchor KLT tracking.

\textbf{Runtime assurance, safe exploration, and evaluation practice.} Simplex runtime assurance~\citep{sha2001simplex}, learned recovery zones~\citep{thananjeyan2021recoveryrl}, and OOD or attack-recovery wrappers~\citep{dash2024specguard,ganai2025fortress} engage a protective backup when a monitor triggers; safe-exploration methods preserve ergodicity via return policies~\citep{moldovan2012safe} or project unsafe actions through a safety layer~\citep{dalal2018safetylayer}. Our module applies this wrapper architecture at the trajectory-reference and feedback-translation boundary without touching low-level actuation. Following rigorous empirical practice~\citep{henderson2018matters,agarwal2021precipice}, we preregister primary comparisons on fresh scenes and evaluate uncertainty with paired scene-cluster bootstrap intervals~\citep{efron1994bootstrap}.

\section{Problem Formulation and Kinematic Analysis}
\label{sec:problem}

A stereo visual-inertial quadrotor follows a horizontal circular route of radius $R = 1.0\,\mathrm{m}$ at nominal speed $v = 0.1925\,\mathrm{m/s}$ (after a $2\,\mathrm{s}$ ramp) over horizon $H = 32\,\mathrm{s}$ without external localization (Appendix~B, \cref{tab:sensing}). At each step, the simulator generates $20\,\mathrm{Hz}$ stereo images and $500\,\mathrm{Hz}$ IMU measurements, passes images to Basalt VIO~\citep{usenko2020basalt} when visible ($v_t = 1$), and withholds them during blackout ($v_t = 0$).

\textbf{Two aliased outage classes.} Each flight experiences one undisclosed outage:
\begin{itemize}
\setlength\itemsep{0.5pt}
\item \emph{Exogenous timed blackout} $\HT(D)$: $v_t = \1[t \notin [t_{\mathrm{on}}, t_{\mathrm{on}}+D)]$; primarily \Tb{} ($[10, 18)\,\mathrm{s}$, $D=8\,\mathrm{s}$). Vision returns solely with elapsed time.
\item \emph{Endogenous position-dependent zone} $\HS(\Zzone)$: $v_t = \1[x_t \notin \Zzone]$; primarily \Sl{}, a fixed $L_z = 1.35\,\mathrm{m}$ arc ($[1.7325, 3.0825]\,\mathrm{m}$, radial half-width $0.50\,\mathrm{m}$, $z \in [-0.5, 3.5]\,\mathrm{m}$). Vision returns only when the vehicle physically exits $\Zzone$.
\end{itemize}

\textbf{Parallel task-completion criteria.} A $500\,\mathrm{Hz}$ safety guard terminates a flight as a hard failure (HF) on tracking divergence $\|p_{\mathrm{true}} - p_{\mathrm{cmd}}\| > 2.0\,\mathrm{m}$ (code~3), altitude $z < 0$ or $z > 3.0\,\mathrm{m}$ (code~4), tilt $>80^\circ$ (code~5), speed $>8.0\,\mathrm{m/s}$ (code~6), angular rate $>35\,\mathrm{rad/s}$ (code~7), or nonfinite state/policy outputs. Let $s_0$ be the arc at nominal outage onset ($1.7325\,\mathrm{m}$) and $W = [31, 32]\,\mathrm{s}$ the final scoring window. Physical completion $\Cphys$ holds when a flight avoids HF, crosses four downstream gates at $s_0 + \{0.25, 0.50, 0.75, 1.0\}\,\mathrm{m}$ before $31\,\mathrm{s}$, stays downstream throughout $W$, and maintains final-window route error $\le 0.30\,\mathrm{m}$. Localization-qualified completion additionally requires visual lock and final-window position error $\le 0.15\,\mathrm{m}$ on either:
\begin{itemize}
\setlength\itemsep{0.5pt}
\item $\Cctrl$ (primary): controller localization state $\ploc = \pvio + \alpha\Delta$ (strictly excluding height-integral trim);
\item $\Cvio$ (raw VIO): uncorrected Basalt position $\pvio$.
\end{itemize}
To reward robust recovery across undisclosed regimes rather than single-regime specialization, each arm $a$ is scored by worst-condition completion:
\begin{equation}
W(a) \;=\; \min\big\{\,\hat{C}_{\mathrm{ctrl}}(a, \Tb{}),\;\hat{C}_{\mathrm{ctrl}}(a, \Sl{})\,\big\}.
\label{eq:worst}
\end{equation}

\textbf{1D kinematic and blind-exposure analysis.} To analyze probe-budget feasibility and blind-exposure decomposition, consider a 1D arc-length model $x_t \in \mathbb{R}$ advancing at constant speed $v > 0$ into $\HT(D)$ or $\HS(\Zzone)$ ($\Zzone = [s_{\mathrm{in}}, s_{\mathrm{out}}]$, $L_z = s_{\mathrm{out}} - s_{\mathrm{in}} > 0$) at $t_{\mathrm{in}} = 0$. Online telemetry exposes only $v_t$, inertial propagation, and prior commands, aliasing $\HT$ and $\HS$ at detection $t_0 = \tau_{\mathrm{det}} > 0$. After supervisor tick delay $\delta \in [0, \Delta t_{\mathrm{tick}}]$, the supervisor commands exact tracking of the minimum-duration quintic velocity transition from $v$ to $0$ under limits $(a_{\max}, j_{\max})$:
\begin{equation}
T_*(v) \;=\; \max\!\left\{\frac{15\,v}{8\,a_{\max}},\; \sqrt{\frac{10\,v}{\sqrt{3}\,j_{\max}}}\right\}.
\label{eq:tstar}
\end{equation}
Blind age $b_t$ advances at unit rate while $v_t=0$ and resets to $0$ upon visual re-lock (requiring clear margin $d_{\mathrm{relock}} \ge 0$ outside $\Zzone$), with any strictly increasing blind-drift proxy $\Phi(b)$.

\begin{figure*}[t]
\centering
\includegraphics[width=\textwidth]{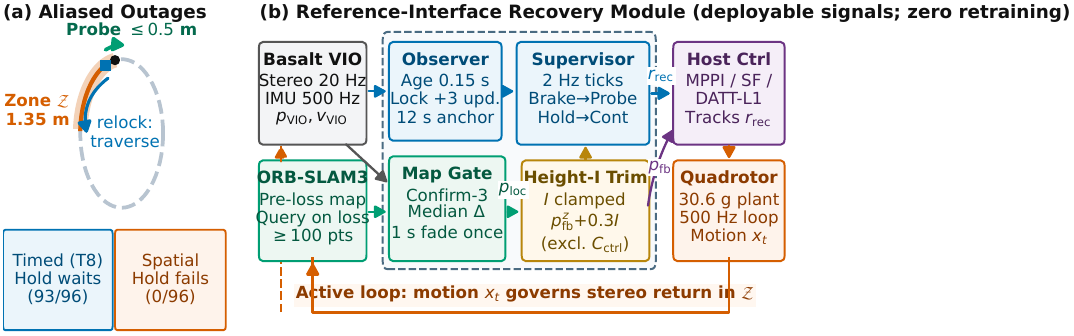}
\caption{Problem structure and portable recovery architecture. (a)~Timed (\Tb{}) and spatial (\Sl{}) outages are observationally aliased at loss onset under online telemetry; a bounded backward probe (0.50\,m planned budget) exits $\Zzone$ to restore KLT tracking and latch \texttt{spatial\_seen}. (b)~The recovery module operates solely on deployable telemetry and references without modifying host controller weights, outputting recovery reference $\rrec$ and feedback position $\pfb$ (with optional support- and consistency-gated ORB-SLAM3 translation $\alpha\Delta$). $\Cctrl$ scores $\ploc$ before height-integral trim; $\Cvio$ scores raw $\pvio$.}
\label{fig:architecture}
\end{figure*}

\begin{lemma}[1D Kinematic Stopping-Depth and Blind-Exposure Bounds]
\label{lem:decision}
In the 1D kinematic model above under exact profile tracking:
\begin{enumerate}
\setlength\itemsep{0.5pt}
\item \textbf{Hold collapse under $\HS$:} If a passive hold policy $\pi_{\mathrm{hold}}$ stops inside $\Zzone$ ($d_{\mathrm{stop}} < L_z$) and holds true position fixed until $v_t=1$, then $v_t \equiv 0$ for all $t \ge t_0$, yielding $\Cctrl(\pi_{\mathrm{hold}}; \HS) = W(\pi_{\mathrm{hold}}) = 0$.
\item \textbf{Stopping depth and rear-boundary exit:} Under pre-brake speed $v$ and braking duration $T_*(v)$, penetration depth inside $\Zzone$ is
\begin{equation}
\begin{aligned}
d_{\mathrm{stop}}(v, \delta) &= v(\tau_{\mathrm{det}} + \delta) + \tfrac{1}{2} v T_*(v) \\
&\le v(\tau_{\mathrm{det}} + \Delta t_{\mathrm{tick}}) + \tfrac{1}{2} v T_*(v).
\end{aligned}
\label{eq:dpen}
\end{equation}
Whenever $d_{\mathrm{stop}}(v, \delta) < L_z$, stopping depth is independent of $L_z$, and a backward probe of budget $L_p > d_{\mathrm{stop}}(v, \delta) + d_{\mathrm{relock}}$ exits $s_{\mathrm{in}}$ and restores $v_t = 1$.
\item \textbf{Two-segment blind-exposure reduction:} Under $\HS(\Zzone)$ with $d_{\mathrm{stop}} < L_z$ and $L_p > d_{\mathrm{stop}} + d_{\mathrm{relock}}$, backward probing splits blindness into a probe segment $\tau_{\mathrm{probe}}$, a visual reset, and a forward crossing $\tau_{\mathrm{trav}} = L_z/v$. By contrast, a non-probing hold-then-continue policy $\pi_{\mathrm{HC}}(T_w)$ ($T_w \ge 0$) traverses $L_z - d_{\mathrm{stop}}$ in a single blind stretch $\tau_{\mathrm{HC}} = \tau_{\mathrm{det}} + \delta + T_*(v) + T_w + (L_z - d_{\mathrm{stop}})/v = \tau_{\mathrm{trav}} + T_w + \tfrac{1}{2} T_*(v) > \tau_{\mathrm{trav}}$. Conditional on $\tau_{\mathrm{probe}} < \tau_{\mathrm{HC}}$, peak blind drift is strictly lower: $\max\{\Phi(\tau_{\mathrm{probe}}), \Phi(\tau_{\mathrm{trav}})\} < \Phi(\tau_{\mathrm{HC}})$. Under $\HT(D)$, images return before fallback continuation whenever $\tau_{\mathrm{probe},T} + T_h \ge D$.
\end{enumerate}
\end{lemma}

Appendix~A proves \cref{lem:decision}. Substituting our frozen constants ($a_{\max}=0.5\,\mathrm{m/s}^2$, $j_{\max}=0.7\,\mathrm{m/s}^3$, $\tau_{\mathrm{det}}=0.15\,\mathrm{s}$, $\Delta t_{\mathrm{tick}}=0.5\,\mathrm{s}$; code uses conservative $5.774 \ge 10/\sqrt{3}$) gives mid-tick depths of $0.20\,\mathrm{m}$ ($1.0\times$) and $0.27\,\mathrm{m}$ ($1.25\times$), closely tracking archived trace medians ($0.21\,\mathrm{m}$, $0.25\,\mathrm{m}$), while 3D empirical maxima ($0.30\,\mathrm{m}$, $0.37\,\mathrm{m}$) remain well inside the $0.50\,\mathrm{m}$ planned probe budget. Worst-tick $d_{\mathrm{stop}}$ matches the $0.50\,\mathrm{m}$ budget near $1.75\times$ speed ($0.337\,\mathrm{m/s}$), whereas the $32\,\mathrm{s}$ circular route hits next-lap re-entry into $\Zzone$ near $1.34\times$ (Appendix~A). Analytical evaluation of a $4\,\mathrm{s}$ hold-then-continue profile $\pi_{\mathrm{HC}}(4\,\mathrm{s})$ yields $11.5$--$12.2\,\mathrm{s}$ of contiguous blindness compared with our module's $7.2\,\mathrm{s}$ median (max $11.2$--$13.9\,\mathrm{s}$).

\textbf{Experimental protocol.} Procedural scenes are admitted solely by healthy native-MPPI completion (RMSE $\le 0.30\,\mathrm{m}$), and all arms in a batch share one seed tuple per scene $\times$ execution. Primary CIs use $10{,}000$ paired scene-cluster percentile bootstrap resamples~\citep{efron1994bootstrap} (secondary trajectory metrics use $2{,}000$; Appendix~F). In EV4 (48 fresh scenes, frozen in \texttt{EV4\_FREEZE.md}), H1 (95\%) and six external worst-$\Cctrl$ contrasts (Bonferroni 99.17\%) form the confirmatory test family, supplemented by a smooth linear sensitivity check in Appendix~E; follow-up batches are analyzed separately alongside EV4.

\begin{table*}[t]
\centering
\scriptsize
\setlength{\tabcolsep}{3.0pt}
\caption{Task completion across confirmatory EV4 (\textbf{A}, 48 fresh scenes $\times$ 2 execs, $n=96$/cell), paired follow-up batches on EV4 scenes/seeds (\textbf{B}), and two-speed EV6 (\textbf{C}, 10 fresh scenes, $n=20$/cell, analyzed separately). Slash notation denotes \Tb{}/\Sl{} in \textbf{A}--\textbf{B} (HF and Delay) and $1.0\times / 1.25\times$ in \textbf{C} (HF sums both outages per speed; Delay shows \Tb{}). ``Mod.'' = supervisor + height-I; DATT-L1 uses supervisor alone. Worst $\Cctrl$ reports 95\% scene-cluster intervals; Delay is median end-of-flight true-progress lag (s) on non-failed runs. Full $\Cvio$ and contrasts appear in \cref{app:tables}.}
\label{tab:main}
\begin{tabular}{llrrcrrr}
\toprule
 & Arm & \Tb{} & \Sl{} & Worst $\Cctrl$ \% [95\%] & Worst $\Cphys$ \% & HF & Delay (s) \\
\midrule
\multirow{10}{*}{\rotatebox{90}{\textbf{A}\; EV4, $n{=}96$}}
 & \textbf{MPPI + mod. + map (full)} & \textbf{93} & \textbf{85} & \textbf{88.5} [81.2, 94.8] & \textbf{90.6} & 1 / 0 & 12.7 / 9.4 \\
 & MPPI + mod., map off & 92 & 70 & 72.9 [63.5, 81.3] & 76.0 & 1 / 6 & 12.6 / 9.4 \\
 & DATT-L1 + sup. (no height-I) & 78 & 67 & 69.8 [60.4, 78.1] & 77.1 & 0 / 0 & 13.9 / 9.5 \\
 & Native MPPI & 70 & 69 & 71.9 [61.5, 79.2] & 76.0 & 7 / 6 & 0.1 / 0.2 \\
 & SimpleFlight (tuned, C4) & 66 & 78 & 68.8 [59.4, 77.1] & 70.8 & 12 / 4 & 0.2 / 0.1 \\
 & DATT-L1 (tuned, C7; H1) & 57 & 67 & 59.4 [49.0, 69.8] & 60.4 & 2 / 3 & 0.2 / 0.1 \\
 & L1-MPPI & 71 & 67 & 69.8 [59.4, 78.1] & 72.9 & 9 / 3 & 0.1 / 0.1 \\
 & LOTF base (retrained) & 13 & 27 & 13.5 [6.2, 21.9] & 14.6 & 12 / 3 & 0.8 / $-0.5$ \\
 & LOTF online & 15 & 15 & 15.6 [8.3, 19.8] & 15.6 & 2 / 7 & $-0.5$ / $-0.3$ \\
 & LOTF online, paused in loss & 11 & 6 & 6.2 [2.1, 10.4] & 6.2 & 9 / 18 & $-0.2$ / $-0.3$ \\
\midrule
\multirow{2}{*}{\textbf{B}}
 & Hover-and-wait (MPPI + height-I) & 93 & 0 & 0.0 & 0.0 & 2 / 13 & 10.4 / 20.5 \\
 & SimpleFlight + mod. (no map) & 94 & 83 & 86.5 & 86.5 & 0 / 4 & 13.0 / 9.3 \\
\midrule
\multirow{5}{*}{\rotatebox{90}{\textbf{C}\; EV6}}
 & Full ($1.0\times / 1.25\times$) & 19 / 20 & 15 / 16 & 75 [55, 95] / 80 [65, 95] & 75 / 80 & 2 / 1 & 12.7 / 12.1 (\Tb{}) \\
 & MPPI + mod., map off & 18 / 20 & 14 / 14 & 70 [50, 90] / 70 [50, 90] & 75 / 70 & 2 / 1 & 12.5 / 12.0 (\Tb{}) \\
 & Native MPPI & 16 / 11 & 14 / 15 & 70 [50, 90] / 55 [40, 70] & 70 / 55 & 4 / 8 & 0.0 / 0.1 (\Tb{}) \\
 & SimpleFlight (tuned, C4) & 15 / 14 & 18 / 19 & 75 [55, 90] / 70 [45, 90] & 80 / 70 & 3 / 5 & 0.1 / 0.1 (\Tb{}) \\
 & \textbf{SimpleFlight + mod. (no map)} & \textbf{20} / \textbf{20} & 18 / 18 & \textbf{90} [75, 100] / \textbf{90} [75, 100] & \textbf{90} / \textbf{90} & 0 / 0 & 13.2 / 12.2 (\Tb{}) \\
\bottomrule
\end{tabular}
\end{table*}

\section{Portable Active Recovery Module}
\label{sec:method}

The recovery module (\cref{fig:architecture}) sits between stereo VIO and an unchanged host tracker, exchanging only deployable telemetry and trajectory references (Appendix~B, \cref{tab:sensing}; Appendix~C, \cref{tab:constants}).

\textbf{Observer.} Visual loss is declared once accepted VIO visual-update age exceeds \texttt{LOSS\_AGE\_S = 0.15\,s} ($\approx 0.11$--$0.13\,\mathrm{s}$ from zone entry at 20\,Hz). Opening an episode from a fully locked state latches the last accepted visual timestamp as the blind-budget anchor (retained across partial re-losses). Visual re-lock requires recovery plus three further accepted visual updates.

\textbf{Supervisor.} Stepping at 50\,Hz on $\pfb$ and $v_{\mathrm{VIO}}$, the supervisor logs breadcrumbs $B$ every $0.02\,\mathrm{m}$ (including during blackout) and executes mode transitions on 2\,Hz ($0.5\,\mathrm{s}$) ticks (\cref{alg:supervisor}). Its nominal mission clock advances \emph{only} in \texttt{NOM} and \texttt{CONT}, pausing while stopped. On loss with \texttt{spatial\_seen} unset, the supervisor executes a quintic brake ($a \le 0.5\,\mathrm{m/s}^2$, $j \le 0.7\,\mathrm{m/s}^3$) and retraces $B$ backward over a $0.50\,\mathrm{m}$ planned budget at $\le 0.25\,\mathrm{m/s}$ (once per episode; the terminal braking ramp adds a short bounded tail, e.g.\ $0.54\,\mathrm{m}$ total in archived logs).

When vision re-locks during backward probing, the supervisor latches \texttt{spatial\_seen} for the flight and brakes to \texttt{HOLD}, so subsequent loss events traverse via \texttt{CONT} rather than re-probing. If the probe completes without vision return (as in \Tb{}), the vehicle holds and falls back to \texttt{CONT} $6\,\mathrm{s}$ later. After re-lock in \texttt{HOLD}, \texttt{SETTLE} waits for speed $\le 0.20\,\mathrm{m/s}$ over $0.5\,\mathrm{s}$ within $1\,\mathrm{m}$ of the route ($>3\,\mathrm{s} \to$ terminal \texttt{STAY}) and rejoins via a quintic blend ($\ge 3\,\mathrm{s}$) to \texttt{NOM}. Re-lock in \texttt{CONT} activates a jump guard ($>0.30\,\mathrm{m}$ offset jump within $0.5\,\mathrm{s} \to$ brake and \texttt{SETTLE}; else \texttt{NOM}). Blind exposure exceeding $12\,\mathrm{s}$ from the anchor enters terminal \texttt{EXHAUSTED}.

\textbf{Height-integral compensation.} On the measured MPPI and LOTF interfaces, the simulated plant produces $\approx 1.2\,\mathrm{m/s}^2$ higher vertical acceleration than nominal hover parameters. After 200 valid 100\,Hz queries, an anti-windup clamped integrator $I \gets \mathrm{clip}(I + 0.01(\ploc^z - \pref^z), \pm 0.1/0.3)$ trims vertical feedback via $\pfb^z = \ploc^z + 0.3I$ ($\pm 0.1\,\mathrm{m}$ cap), strictly excluded from $\Cctrl$.

\textbf{Optional prior-map translation branch.} After the first visual loss, stereo ORB-SLAM3~\citep{campos2021orbslam3} queries the retained pre-outage map (Appendix~C, \cref{alg:map}). Frames with valid anchor, pose, stereo pair, and $\ge 100$ pre-outage map points produce translation candidate $\delta = \pmap - \pvio$; three consecutive candidates with adjacent $\|\delta_i - \delta_j\| \le 0.05\,\mathrm{m}$ confirm coordinate-wise median $\Delta$ (invalid frames reset the window while preserving $\Delta$). On the \emph{first} eligible confirmation, $\alpha$ ramps linearly $0 \to 1$ over $1.0\,\mathrm{s}$ at 10\,Hz ($\ploc = \pvio + \alpha\Delta$). Subsequent confirmations update $\Delta$ directly without restarting the fade (applying at $\alpha = 1$ once the initial ramp finishes).

\textbf{Interface contract and host integration.} Portability requires only a host controller capable of tracking a position/velocity reference $\rrec$ from an external position estimate without weight updates. We evaluate zero-retraining integration across three distinct hosts: MPPI~\citep{huang2023datt} (supervisor + height-I, with and without the prior-map branch), SimpleFlight~\citep{chen2025simpleflight} (supervisor + height-I), and DATT-L1~\citep{huang2023datt} (supervisor alone on raw $\pvio$).

\begin{algorithm}[t]
\caption{Supervisor decision logic (2\,Hz ticks; 50\,Hz tracking)}
\label{alg:supervisor}
\begin{algorithmic}[1]
\scriptsize
\REQUIRE Observer (\texttt{lost}, \texttt{locked}, \texttt{anchor}), $\pfb$, breadcrumbs $B$, \texttt{mode}
\STATE Mission clock $s$ advances \textbf{only} in $\{\texttt{NOM}, \texttt{CONT}\}$
\IF{\texttt{lost} \AND $\texttt{mode} = \texttt{NOM}$} \STATE open episode; $\texttt{mode} \gets \texttt{CONT}$ \ENDIF
\IF{\texttt{lost} \AND $\texttt{mode} \in \{\texttt{CONT}, \texttt{HOLD}\}$}
  \IF{\texttt{spatial\_seen}} \STATE \texttt{CONT} along route
  \ELSIF{probe unused in episode} \STATE brake; \texttt{PROBE} along reversed $B$ (0.5\,m plan)
  \ELSIF{probe ended $\ge 6$\,s ago} \STATE \texttt{CONT} along route
  \ELSE \STATE \texttt{HOLD}
  \ENDIF
\ELSIF{\texttt{locked} \AND $\texttt{mode} = \texttt{HOLD}$ \AND episode open}
  \STATE \texttt{SETTLE} ($\le 3$\,s, else \texttt{STAY}) $\to$ quintic \texttt{BLEND} ($\ge 3$\,s) $\to$ \texttt{NOM}
\ENDIF
\STATE \textbf{Events:} relock in \texttt{PROBE} $\Rightarrow$ \texttt{spatial\_seen}, brake, \texttt{HOLD}; relock in \texttt{CONT} $\Rightarrow$ jump guard ($>0.3$\,m/$0.5$\,s $\to$ \texttt{SETTLE}, else \texttt{NOM}); blind $>12$\,s $\Rightarrow$ \texttt{EXHAUSTED}
\end{algorithmic}
\end{algorithm}

\section{Experiments and Mechanistic Diagnostics}
\label{sec:experiments}

We evaluate the recovery module across six stages: (i)~preregistered confirmation on 48 fresh scenes (EV4, 1{,}920 flights), (ii)~component comparisons across recovery configurations illustrated on archived closed-loop timelines, (iii)~archived-trace mechanistic diagnostics isolating visual-inertial re-acquisition asymmetry, (iv)~zero-retraining transfer across three host controllers, (v)~transfer across two cruise speeds (EV6), and (vi)~rule-selection analysis, bias--drift behavior on real EuRoC data, and operational tradeoffs.

\textbf{Baselines and setup.} EV4 compares ten arms (\cref{tab:main}A): full MPPI (module + map), map-off MPPI, DATT-L1 + supervisor (no height-I), and seven external baselines across four controller families---native MPPI~\citep{williams2017mppi,huang2023datt}, official SimpleFlight~\citep{chen2025simpleflight}, official DATT-L1~\citep{huang2023datt}, L1-MPPI~\citep{hovakimyan2010l1,wu2022l1quad}, and official-code retrained LOTF~\citep{pan2026lotf} (frozen base policy, re-implemented online adaptation, and paused during loss). SimpleFlight and DATT-L1 each screened up to eight deployment configurations on eight separate development scenes, freezing the best by worst-condition $\Cctrl$ (\cref{app:baselines}; DATT-L1 C7 defines H1). Our recovery constants and map gate were frozen prior to an earlier disjoint 48-scene confirmation EV3 (\cref{app:tables}).

\begin{figure*}[t]
\centering
\includegraphics[width=\textwidth]{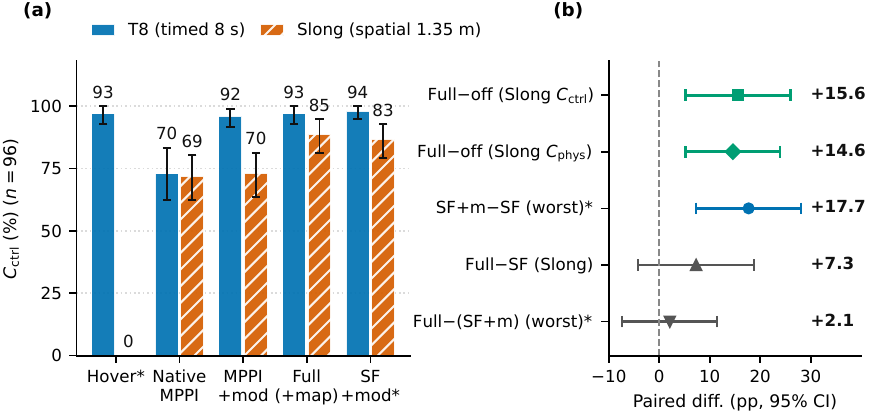}
\caption{Evaluated recovery configurations. (a)~Per-condition $\Cctrl$ completions ($n=96$) with exact 10{,}000-resample 95\% scene-cluster bootstrap intervals. Asterisks (*) mark paired follow-up batches on EV4 scenes and seeds (HLD5 hover-and-wait on MPPI + height-I; PSF4 SimpleFlight + module without map). (b)~Paired 95\% scene-cluster contrasts (pp); confirmatory tests use worst-condition $\Cctrl$ (\cref{tab:main}).}
\label{fig:mechanism}
\end{figure*}

\subsection{Confirmatory Benchmark (EV4)}
\label{sec:confirmatory}

In EV4 (\cref{tab:main}A; 192/192 complete contexts, zero unscored runs; prior disjoint confirmation EV3 in \cref{app:tables}), the full MPPI configuration completes 93/96 \Tb{} and 85/96 \Sl{}, attaining $W = 88.5\%$ [81.2, 94.8]. The preregistered primary contrast H1 (full minus DATT-L1 C7 on worst-condition $\Cctrl$) reaches $+29.2$\pp{} [17.7, 40.6] (exec~6: $+25.0$\pp{}; exec~7: $+33.3$\pp{}), comparing full's \Sl{} rate (88.5\%) against DATT-L1's \Tb{} rate (59.4\%). All six prespecified external-comparator contrasts on worst-condition $\Cctrl$ remain strictly above zero under Bonferroni-adjusted 99.17\% intervals: native MPPI $+16.7$\pp{} [4.2, 32.3], SimpleFlight $+19.8$\pp{} [3.1, 36.5], DATT-L1 $+29.2$\pp{} [14.6, 44.8], L1-MPPI $+18.8$\pp{} [5.2, 34.4], LOTF base $+75.0$\pp{} [62.5, 87.5], and LOTF online $+72.9$\pp{} [62.5, 85.8] (LOTF paused sits outside the prespecified family). A smooth linear sensitivity check avoiding the nonsmooth minimum confirms strictly positive lower bounds for all six comparators (\cref{app:tables}), and physical completion $\Cphys$ agrees throughout (full 90.6\% [84.4, 95.8]).

External baselines fail through distinct mechanisms. Schedule-keeping trackers incur 7--12 \Tb{} hard failures while flying blind on drifting inertial states; DATT-L1 avoids crashes (2/3 HF) but re-converges slowly after vision returns; LOTF lacks a time/phase input and exhibits offline-verified radial update variance, with online adaptation lowering \Sl{} by $-12.5$\pp{} [$-19.8$, $-5.2$] relative to its base policy and pausing during loss scoring lower still (online minus paused: $+9.4$\pp{} [2.1, 16.7] in \Sl{}; $+9.4$\pp{} [1.0, 14.6] worst-condition; \cref{app:baselines}).

\subsection{Component Comparisons Across Recovery Modes}
\label{sec:mechanism}

\begin{figure*}[t]
\centering
\includegraphics[width=0.98\textwidth]{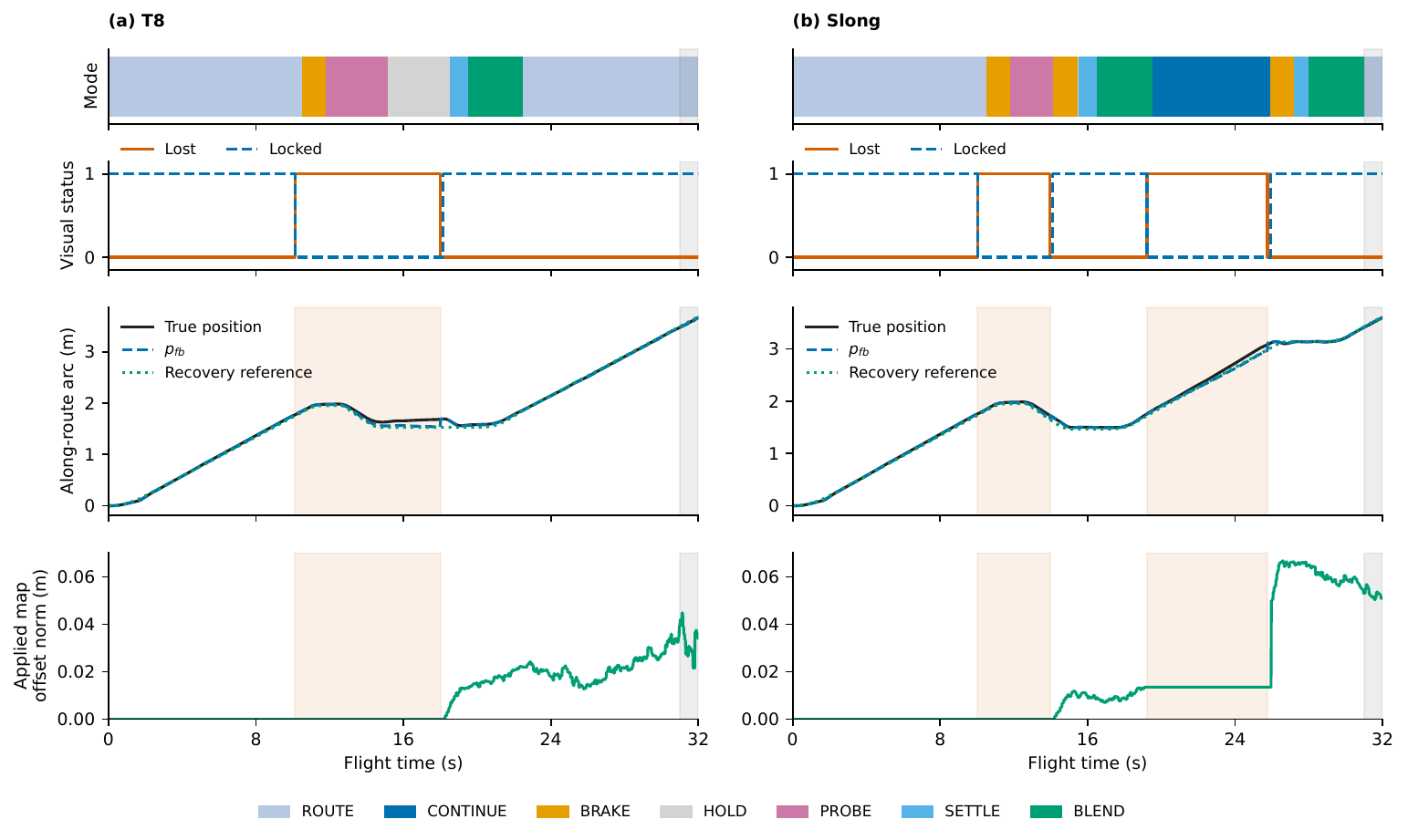}
\caption{Archived EV4 full-system flight timelines (first EV4 scene, execution~6; fixed index). Orange shading: observed visual loss; grey shading: final scoring window $[31, 32]\,\mathrm{s}$. (a)~\Tb{}: brake, backward probe, hold until images return, settle, quintic blend, and one-time 1\,s map fade. (b)~\Sl{}: backward probe exits $\Zzone$ and restores KLT lock (latching \texttt{spatial\_seen}); on re-entry, the vehicle traverses $\Zzone$ in \texttt{CONT}, re-locks on the far side, applies a step map update at $\alpha = 1$, and completes the task.}
\label{fig:timeline}
\end{figure*}

\cref{fig:mechanism} contrasts the evaluated configurations, and \cref{fig:timeline} illustrates both complete recovery sequences directly on archived EV4 flights. Passive hover-and-wait (HLD5, MPPI + height-I without probing or timed continuation) matches the full system on \Tb{} at 93/96 ($0.0$\pp{} [$-4.2$, $+4.2$], 2/2 discordant), demonstrating that stopping suffices under timed blackouts, yet collapses to 0/96 in \Sl{} (13 HF). Enabling the full probe-and-continue supervisor without a map restores vision during probing in 93--96/96 \Sl{} flights (median $\approx 3.4$\,s) and lifts \Sl{} completion from 0 to 70/96 ($+72.9$\pp{} [63.5, 81.2]).

On MPPI, adding the gated prior-map branch lifts \Sl{} completion from 70 to 85/96 ($+15.6$\pp{} [5.2, 26.0] in $\Cctrl$; $+14.6$\pp{} [5.2, 24.0] in $\Cphys$) while aggregate $\Cvio$ stays at 70/96 (\cref{tab:ev4full}), confirming that the map acts strictly through controller localization $\ploc$. Whereas forward continuation suffices on many spatial crossings taken in isolation (native MPPI 69/96, SimpleFlight 78/96), maintaining high completion across both undisclosed regimes requires pairing timed-outage stopping with active spatial recovery and re-acquisition.

\subsection{Mechanistic Diagnostics: Re-Acquisition Asymmetry}
\label{sec:diagnostics}

\begin{figure*}[t]
\centering
\includegraphics[width=\textwidth]{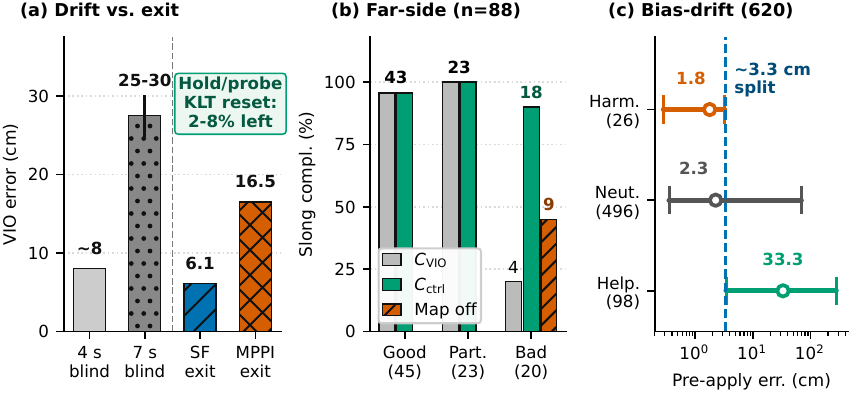}
\caption{Archived-trace mechanistic diagnostics. (a)~Absolute VIO position error (cm): blind drift grows similarly across actions and hosts ($\approx 8$\,cm at 4\,s, $\le 1$\,cm MPPI vs.\ SF gap despite $4.4\times$ rate excitation; 25--30\,cm at 7\,s); after hold/probe, KLT retains only 2--8\% of blind error (callout), whereas \Sl{} far-side exits leave higher median error on MPPI + module (16.5\,cm) than SF + module (6.1\,cm). (b)~EV4 \Sl{} full-system runs with a classified far-side transition ($n=88$ of 96; 8 unclassified): on the 20 bad-class flights, completion rises from 4/20 ($\Cvio$) to 18/20 ($\Cctrl$), vs.\ 9/20 for paired map-off ($+45$\pp{} [11, 71]). (c)~Offline pre-apply VIO error across 620 applied map segments (EV3+EV4+SWP5): all 26 harmful ($\le 3.25$\,cm) and all 98 helpful ($\ge 3.44$\,cm) separate cleanly at $3.3$\,cm (496 neutral span both sides).}
\label{fig:diagnostics}
\end{figure*}

Archived EV4 trace diagnostics (\cref{fig:diagnostics}a) overturn the intuition that stopping works by suppressing blind IMU drift. Open-loop inertial drift grows quadratically ($\approx 0.08$\,m at 4\,s, $0.25$--$0.30$\,m at 7\,s) and differs by $\le 1$\,cm at 4\,s between MPPI + module and SimpleFlight + module despite $4.4\times$ higher body-rate excitation on MPPI (only 2/8 paired comparisons exclude zero). Instead, outcomes separate at vision return. After hold or backward probing back into seen texture, Basalt's frame-to-frame KLT optical flow re-anchors immediately, leaving just 2--8\% of accumulated blind error. After forward traversal across unseen \Sl{} texture, far-side re-acquisition yields 88--100\% raw-VIO completion on clean exits versus 5/21 (MPPI + module) and 1/9 (SimpleFlight + module) on bad exits. Bad far-side re-acquisitions occur in 13--29\% of MPPI and DATT-L1 flights versus 6--11\% on SimpleFlight (median error at loss-flag clear $0.165$ vs.\ $0.061$\,m across 96 paired \Sl{} runs; paired mean difference $+0.18$\,m [0.06, 0.30]).

Joining archived far-side re-acquisition classes (classified on EV4 \Sl{} traces) with outcomes (\cref{fig:diagnostics}b; 88/96 scorable runs, 8 unclassified kept in all 96-run totals) pinpoints where the map branch rescues performance. Good ($n=45$) and partial ($n=23$) exits complete 43/45 and 23/23 identically across $\Cvio$, $\Cctrl$, and $\Cphys$. On the 20 bad-class flights, prior-map translation raises same-flight completion from 4/20 ($\Cvio$) to 18/20 ($\Cctrl$ and $\Cphys$, capturing 14/15 $\Cctrl$/$\Cvio$-discordant flights), compared with 9/20 for paired map-off runs on those same 20 contexts ($+45$\pp{} [11, 71]).

\subsection{Transfer Across Unchanged Host Controllers}
\label{sec:transfer}

\begin{table}[t]
\centering
\scriptsize
\setlength{\tabcolsep}{1.8pt}
\caption{Zero-retraining host transfer: host alone $\to$ with recovery components ($\Cctrl$ counts, worst \%, and combined hard failures across \emph{both} outages: $n_{\mathrm{HF}}=192$ in EV4/PSF4, $40$ in EV6). *Paired batch PSF4 on EV4 scenes/seeds.}
\label{tab:transfer}
\begin{tabular}{lcccc}
\toprule
Host + added ($n$) & \Tb{} & \Sl{} & Worst \% & HF \\
\midrule
MPPI + mod. (96) & 70$\to$92 & 69$\to$70 & 71.9$\to$72.9 & 13$\to$7 \\
MPPI + mod. + map (96) & 70$\to$93 & 69$\to$85 & 71.9$\to$88.5 & 13$\to$1 \\
SF + mod. (96)* & 66$\to$94 & 78$\to$83 & 68.8$\to$86.5 & 16$\to$4 \\
DATT-L1 + sup. (96) & 57$\to$78 & 67$\to$67 & 59.4$\to$69.8 & 5$\to$0 \\
SF + mod. $1.0\times$ (20) & 15$\to$20 & 18$\to$18 & 75.0$\to$90.0 & 3$\to$0 \\
SF + mod. $1.25\times$ (20) & 14$\to$20 & 19$\to$18 & 70.0$\to$90.0 & 5$\to$0 \\
\bottomrule
\end{tabular}
\end{table}

\cref{tab:transfer} summarizes zero-retraining portability across three distinct host controllers. Attaching the recovery components raises \Tb{} completion on every host (MPPI 70$\to$92, SimpleFlight 66$\to$94, DATT-L1 57$\to$78 of 96) and cuts combined hard failures across both outages ($192$ flights/arm: $13\to 7$, $16\to 4$, $5\to 0$; and $13\to 1$ with the map on MPPI). Host-level worst-condition outcomes align directly with each host's far-side re-acquisition profile:
\begin{itemize}
\setlength\itemsep{1pt}
\item \textbf{SimpleFlight} (PSF4, map off): cleaner far-side exits enable the recovery module to raise worst-condition $\Cctrl$ by $+17.7$\pp{} [7.3, 28.1] to 86.5\% (\Tb{} $+29.2$\pp{} [20.8, 37.5]; \Sl{} $+5.2$\pp{} [$-5.2$, $+15.6$]), closely matching full MPPI (178 vs.\ 177/192 pooled; worst-condition gap $+2.1$\pp{} [$-7.3$, $+11.5$]).
\item \textbf{DATT-L1} (supervisor alone): \Tb{} climbs $+21.9$\pp{} [10.4, 33.3] and worst $\Cphys$ climbs $+16.7$\pp{} [5.2, 26.0] with zero hard failures (worst $\Cctrl$ $+10.4$\pp{} [$-2.1$, $+22.9$], with 9 \Tb{} deadline misses).
\item \textbf{MPPI}: higher far-side re-acquisition error holds map-off worst $\Cctrl$ near native MPPI ($+1.0$\pp{} [$-8.3$, $+13.5$], 71.9\%$\to$72.9\%) until the prior-map branch lifts \Sl{} from 70 to 85/96 (88.5\%).
\end{itemize}

\subsection{Two-Speed Generalization (EV6)}
\label{sec:speed}

\begin{figure*}[t]
\centering
\includegraphics[width=\textwidth]{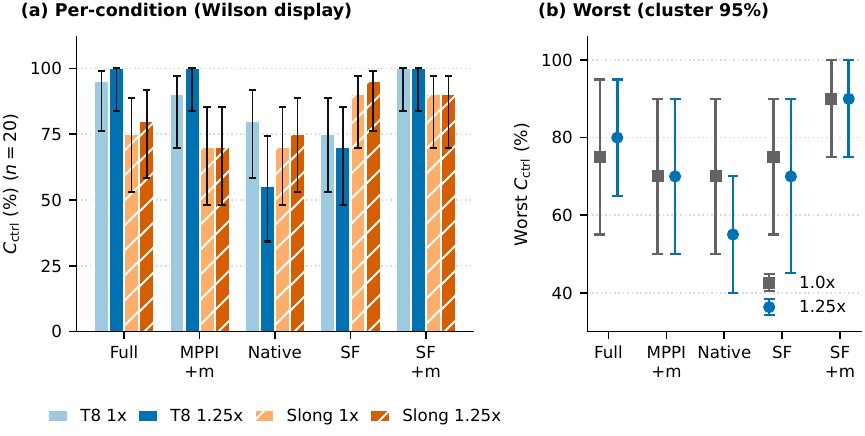}
\caption{EV6 transfer across two cruise speeds on ten fresh scenes ($n=20$ per cell). (a)~Per-condition $\Cctrl$ completion with 95\% Wilson display intervals. (b)~Worst-condition $\Cctrl$ with 95\% scene-cluster bootstrap intervals from \texttt{EV6\_readout.json}.}
\label{fig:speed}
\end{figure*}

EV6 tests five frozen arms on ten fresh scenes at $1.0\times$ ($0.1925\,\mathrm{m/s}$) and $1.25\times$ ($0.240625\,\mathrm{m/s}$; 400 outage runs + 50 healthy $1.25\times$ runs all 10/10; \cref{tab:main}C, \cref{fig:speed}). Faster flight expands blind \Tb{} travel from 1.6 to 1.9\,m for schedule-keeping controllers, dropping native MPPI to 55.0\% worst $\Cctrl$ (7 \Tb{} HF). All three supervised arms complete 20/20 \Tb{} at $1.25\times$; full MPPI reaches 80.0\% [65.0, 95.0] worst $\Cctrl$ ($+25.0$\pp{} [$+5.0$, $+50.0$] over native MPPI; 3/80 HF), and SimpleFlight + module achieves 90.0\% [75.0, 100.0] at both speeds with 0/80 HF. Within-arm speed shifts and map increments remain within 10-scene sampling intervals, and speeds beyond $1.34\times$ would re-enter $\Zzone$ on the second lap inside 32\,s (\cref{app:proof}).

\subsection{Rule Selection, Bias--Drift Tradeoff, and Costs}
\label{sec:costs}

\begin{figure}[t]
\centering
\includegraphics[width=\columnwidth]{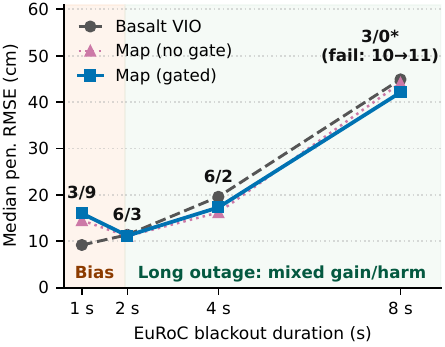}
\caption{Open-loop real-data EuRoC MAV replay across 72 held-out blackout windows (18 per duration; 7 unseen sequences). Labels show windows where the gated map lowers vs.\ raises penalized 10\,s RMSE relative to raw Basalt VIO; *at 8\,s, three improve and zero worsen among seven jointly scored windows while total failures move from 10/18 to 11/18.}
\label{fig:euroc_main}
\end{figure}

\textbf{Rule-selection screening (\cref{app:learning}, \cref{tab:rlp}).} Three pre-freeze screening stages evaluate learned supervision over the same deployable observation space. On development scenes (G1, 4 scenes), a per-context oracle over five fixed rules reached mean terminal \emph{utility} $0.849$ vs.\ $0.840$ for the best single rule ($0.009$ headroom against a $0.10$ launch gate). On 8 fresh validation scenes (RLP, 96 contexts, single training seed), behavior-cloned $\pi_0$ already reached 93/96, PPO~\citep{schulman2017ppo} matched 93/96 with 0/787 greedy decision changes on $\pi_0$'s fixed validation histories (final-per-update $\mathrm{KL}_{k3} \le 0.005$), clipped-LOO reached 94/96, and frozen RPV3 reached 95/96. With near-ceiling validation headroom in this observation space, the deterministic supervisor captures the achievable gain cleanly.

\textbf{Bias--drift crossover and EuRoC replay (\cref{fig:euroc_main}, \cref{app:euroc}, \cref{tab:euroc}).} Across 620 applied map segments in EV3, EV4, and SWP5 (\cref{fig:diagnostics}c), 98 reduced controller error by $>5$\,cm at 3\,s, 26 increased error by $>5$\,cm, and 496 were neutral, with three verified map-on non-completions paired with map-off completion (two in EV4 \Tb{}, one in SWP5 \Ss{}). Offline pre-apply VIO error separates all 26 harmful segments ($\le 3.25$\,cm, med $1.79$\,cm) from all 98 helpful ones ($\ge 3.44$\,cm, med $33.3$\,cm) at $3.3$\,cm: map correction helps when VIO drift exceeds alignment bias and adds bias when KLT has already reset error. Open-loop EuRoC replay~\citep{burri2016euroc} (72 held-out windows, penalized 10\,s RMSE) reproduces this exact crossover: map alignment adds bias after 1\,s blackouts ($0.092\to 0.160$\,m; 3 lower / 9 higher) but reduces error after 4\,s ($0.196\to 0.173$\,m; 6 lower / 2 higher) and on jointly scored 8\,s windows (3 lower / 0 higher, while total 8\,s failures move from 10/18 to 11/18 on \texttt{V2\_03} $0.44\to 1.04$\,m).

\textbf{Nominal no-harm, schedule tradeoff, and repeatability.} Under healthy flight and mild outages (SWP5: healthy, 4\,s timed, 0.58\,m spatial; 864 runs, \cref{app:tables}), all six tested arms (excluding LOTF) complete 47--48/48 runs with zero HF, and the supervisor stays silent in healthy flight (0.06\,s lag). Under severe outages, active recovery trades 9--14\,s of end-of-flight progress lag (12.7\,s \Tb{} / 9.4\,s \Sl{} on full MPPI, raising time-indexed tracking error by 0.89--1.06\,m) for large completion and safety gains while lowering route RMSE, maximum deviation, post-return tilt, and ATE (\cref{app:metrics}). Repeated commands match $\Cctrl$ in 115/120 audited flights (AUD4, median continuous diff $0$--$13$\,mm).

\section{Discussion and Conclusion}
\label{sec:conclusion}

\textbf{Scope and open directions.} Our closed-loop evaluation examines a circular trajectory at two speeds ($0.193$ and $0.241\,\mathrm{m/s}$) under clean stereo frame drops, zero turn-on IMU bias, and workstation execution ($\approx 18$\,ms MPPI and $\approx 76$\,ms/frame ORB-SLAM3 under concurrent load). Extending this framework naturally points to four concrete next steps: (1)~matched timer-only wait-then-continue ablations and probabilistic outage-regime inference that replaces the binary \texttt{spatial\_seen} latch when timed blackouts expire mid-probe; (2)~online drift-conditioned map gating with continuous slew-rate limiting on all post-fade updates, turning the offline $3.3\,\mathrm{cm}$ bias--drift diagnostic into a real-time safeguard against short-outage degradation (as seen on 1\,s EuRoC windows and three verified simulation non-completions); (3)~evaluation across obstacle-rich non-circular routes, partial visual degradation, and physical flights with onboard compute; and (4)~multi-seed learned supervision once broader route and outage distributions create headroom above deterministic rules.

\textbf{Conclusion.} When visual outages can depend on spatial position, task completion under VIO becomes an active-observability problem rather than pure trajectory tracking. Our portable reference-interface recovery module achieves 88.5\% worst-condition completion on a preregistered 48-scene benchmark---exceeding all six external baselines after Bonferroni correction---and transfers without retraining across three distinct flight controllers, lifting SimpleFlight to 86.5\% in EV4 and 90\% across two speeds in EV6 with zero hard failures. Equally important, trace diagnostics shift the mechanistic picture from open-loop IMU drift suppression to visual-inertial re-acquisition asymmetry and a bias--drift map crossover confirmed on real EuRoC data, providing a principled empirical foundation for active recovery under perceptual blackouts.

\FloatBarrier
\section*{Impact Statement}
This work develops recovery logic that helps visual-inertial quadrotors finish planned trajectories despite multi-second camera blackouts, supporting safer autonomous inspection, mapping, and indoor navigation. Because recovery maneuvers brake, briefly retrace, and extend flight duration by roughly 9--14\,s across supervised hosts, any real-world deployment requires onboard obstacle avoidance, physical flight validation, and compliance with local airspace and privacy regulations. We identify no societal risks specific to this study beyond those associated with autonomous aerial robotics generally.

\bibliography{references}

@inproceedings{huang2023datt,
  author    = {Huang, Kevin and Rana, Rwik and Spitzer, Alexander and Shi, Guanya and Boots, Byron},
  title     = {{DATT}: Deep Adaptive Trajectory Tracking for Quadrotor Control},
  booktitle = {Proceedings of The 7th Conference on Robot Learning},
  series    = {Proceedings of Machine Learning Research},
  volume    = {229},
  pages     = {326--340},
  publisher = {PMLR},
  year      = {2023},
  url       = {https://proceedings.mlr.press/v229/huang23a.html}
}

@article{chen2025simpleflight,
  author  = {Chen, Jiayu and Yu, Chao and Xie, Yuqing and Gao, Feng and Chen, Yinuo and Yu, Shu'ang and Tang, Wenhao and Ji, Shilong and Mu, Mo and Wu, Yi and Yang, Huazhong and Wang, Yu},
  title   = {What Matters in Learning a Zero-Shot Sim-to-Real {RL} Policy for Quadrotor Control? {A} Comprehensive Study},
  journal = {IEEE Robotics and Automation Letters},
  volume  = {10},
  number  = {7},
  pages   = {7134--7141},
  year    = {2025},
  doi     = {10.1109/lra.2025.3575011}
}

@article{pan2026lotf,
  author  = {Pan, Jiahe and Xing, Jiaxu and Reiter, Rudolf and Zhai, Yifan and Aljalbout, Elie and Scaramuzza, Davide},
  title   = {Learning on the Fly: Rapid Policy Adaptation via Differentiable Simulation},
  journal = {IEEE Robotics and Automation Letters},
  volume  = {11},
  number  = {3},
  pages   = {3542--3549},
  year    = {2026},
  doi     = {10.1109/lra.2026.3655300}
}

@inproceedings{williams2017mppi,
  author    = {Williams, Grady and Wagener, Nolan and Goldfain, Brian and Drews, Paul and Rehg, James M. and Boots, Byron and Theodorou, Evangelos A.},
  title     = {Information Theoretic {MPC} for Model-Based Reinforcement Learning},
  booktitle = {2017 IEEE International Conference on Robotics and Automation (ICRA)},
  pages     = {1714--1721},
  year      = {2017},
  doi       = {10.1109/icra.2017.7989202}
}

@book{hovakimyan2010l1,
  author    = {Hovakimyan, Naira and Cao, Chengyu},
  title     = {{$\mathcal{L}_1$} Adaptive Control Theory: Guaranteed Robustness with Fast Adaptation},
  publisher = {Society for Industrial and Applied Mathematics},
  year      = {2010},
  doi       = {10.1137/1.9780898719376}
}

@inproceedings{wu2022l1quad,
  author    = {Wu, Zhuohuan and Cheng, Sheng and Ackerman, Kasey A. and Gahlawat, Aditya and Lakshmanan, Arun and Zhao, Pan and Hovakimyan, Naira},
  title     = {{$\mathcal{L}_1$} Adaptive Augmentation for Geometric Tracking Control of Quadrotors},
  booktitle = {2022 International Conference on Robotics and Automation (ICRA)},
  pages     = {1329--1336},
  year      = {2022},
  doi       = {10.1109/icra46639.2022.9811946}
}

@article{oconnell2022neuralfly,
  author  = {O'Connell, Michael and Shi, Guanya and Shi, Xichen and Azizzadenesheli, Kamyar and Anandkumar, Anima and Yue, Yisong and Chung, Soon-Jo},
  title   = {{Neural-Fly} Enables Rapid Learning for Agile Flight in Strong Winds},
  journal = {Science Robotics},
  volume  = {7},
  number  = {66},
  pages   = {eabm6597},
  year    = {2022},
  doi     = {10.1126/scirobotics.abm6597}
}

@article{kaufmann2023champion,
  author  = {Kaufmann, Elia and Bauersfeld, Leonard and Loquercio, Antonio and M{\"u}ller, Matthias and Koltun, Vladlen and Scaramuzza, Davide},
  title   = {Champion-Level Drone Racing Using Deep Reinforcement Learning},
  journal = {Nature},
  volume  = {620},
  number  = {7976},
  pages   = {982--987},
  year    = {2023},
  doi     = {10.1038/s41586-023-06419-4}
}

@article{loquercio2021highspeed,
  author  = {Loquercio, Antonio and Kaufmann, Elia and Ranftl, Ren{\'e} and M{\"u}ller, Matthias and Koltun, Vladlen and Scaramuzza, Davide},
  title   = {Learning High-Speed Flight in the Wild},
  journal = {Science Robotics},
  volume  = {6},
  number  = {59},
  pages   = {eabg5810},
  year    = {2021},
  doi     = {10.1126/scirobotics.abg5810}
}

@article{song2023racing,
  author  = {Song, Yunlong and Romero, Angel and M{\"u}ller, Matthias and Koltun, Vladlen and Scaramuzza, Davide},
  title   = {Reaching the Limit in Autonomous Racing: Optimal Control Versus Reinforcement Learning},
  journal = {Science Robotics},
  volume  = {8},
  number  = {82},
  pages   = {eadg1462},
  year    = {2023},
  doi     = {10.1126/scirobotics.adg1462}
}

@article{eschmann2024l2f,
  author  = {Eschmann, Jonas and Albani, Dario and Loianno, Giuseppe},
  title   = {Learning to Fly in Seconds},
  journal = {IEEE Robotics and Automation Letters},
  volume  = {9},
  number  = {7},
  pages   = {6336--6343},
  year    = {2024},
  doi     = {10.1109/lra.2024.3396025}
}

@inproceedings{todorov2012mujoco,
  author    = {Todorov, Emanuel and Erez, Tom and Tassa, Yuval},
  title     = {{MuJoCo}: A Physics Engine for Model-Based Control},
  booktitle = {2012 IEEE/RSJ International Conference on Intelligent Robots and Systems},
  pages     = {5026--5033},
  year      = {2012},
  doi       = {10.1109/iros.2012.6386109}
}

@misc{engelsman2026certainty,
  author        = {Engelsman, Daniel and Klein, Itzik},
  title         = {Revisiting Certainty Equivalence: The Structural Coupling Between Estimation and Control in Underactuated Nonlinear Systems},
  howpublished  = {arXiv preprint arXiv:2607.07276},
  year          = {2026},
  eprint        = {2607.07276},
  archivePrefix = {arXiv},
  url           = {https://arxiv.org/abs/2607.07276}
}

@article{usenko2020basalt,
  author  = {Usenko, Vladyslav and Demmel, Nikolaus and Schubert, David and St{\"u}ckler, J{\"o}rg and Cremers, Daniel},
  title   = {Visual-Inertial Mapping With Non-Linear Factor Recovery},
  journal = {IEEE Robotics and Automation Letters},
  volume  = {5},
  number  = {2},
  pages   = {422--429},
  year    = {2020},
  doi     = {10.1109/lra.2019.2961227}
}

@article{campos2021orbslam3,
  author  = {Campos, Carlos and Elvira, Richard and G{\'o}mez Rodr{\'\i}guez, Juan J. and Montiel, Jos{\'e} M. M. and Tard{\'o}s, Juan D.},
  title   = {{ORB-SLAM3}: An Accurate Open-Source Library for Visual, Visual--Inertial, and Multimap {SLAM}},
  journal = {IEEE Transactions on Robotics},
  volume  = {37},
  number  = {6},
  pages   = {1874--1890},
  year    = {2021},
  doi     = {10.1109/tro.2021.3075644}
}

@inproceedings{geneva2020openvins,
  author    = {Geneva, Patrick and Eckenhoff, Kevin and Lee, Woosik and Yang, Yulin and Huang, Guoquan},
  title     = {{OpenVINS}: A Research Platform for Visual-Inertial Estimation},
  booktitle = {2020 IEEE International Conference on Robotics and Automation (ICRA)},
  pages     = {4666--4672},
  year      = {2020},
  doi       = {10.1109/icra40945.2020.9196524}
}

@article{qin2018vinsmono,
  author  = {Qin, Tong and Li, Peiliang and Shen, Shaojie},
  title   = {{VINS-Mono}: A Robust and Versatile Monocular Visual-Inertial State Estimator},
  journal = {IEEE Transactions on Robotics},
  volume  = {34},
  number  = {4},
  pages   = {1004--1020},
  year    = {2018},
  doi     = {10.1109/tro.2018.2853729}
}

@inproceedings{mourikis2007msckf,
  author    = {Mourikis, Anastasios I. and Roumeliotis, Stergios I.},
  title     = {A Multi-State Constraint {Kalman} Filter for Vision-aided Inertial Navigation},
  booktitle = {Proceedings 2007 IEEE International Conference on Robotics and Automation},
  pages     = {3565--3572},
  year      = {2007},
  doi       = {10.1109/robot.2007.364024}
}

@article{forster2017preintegration,
  author  = {Forster, Christian and Carlone, Luca and Dellaert, Frank and Scaramuzza, Davide},
  title   = {On-Manifold Preintegration for Real-Time Visual--Inertial Odometry},
  journal = {IEEE Transactions on Robotics},
  volume  = {33},
  number  = {1},
  pages   = {1--21},
  year    = {2017},
  doi     = {10.1109/tro.2016.2597321}
}

@article{liu2020tlio,
  author  = {Liu, Wenxin and Caruso, David and Ilg, Eddy and Dong, Jing and Mourikis, Anastasios I. and Daniilidis, Kostas and Kumar, Vijay and Engel, Jakob},
  title   = {{TLIO}: Tight Learned Inertial Odometry},
  journal = {IEEE Robotics and Automation Letters},
  volume  = {5},
  number  = {4},
  pages   = {5653--5660},
  year    = {2020},
  doi     = {10.1109/lra.2020.3007421}
}

@article{cioffi2023imo,
  author  = {Cioffi, Giovanni and Bauersfeld, Leonard and Kaufmann, Elia and Scaramuzza, Davide},
  title   = {Learned Inertial Odometry for Autonomous Drone Racing},
  journal = {IEEE Robotics and Automation Letters},
  volume  = {8},
  number  = {5},
  pages   = {2684--2691},
  year    = {2023},
  doi     = {10.1109/lra.2023.3252342}
}

@inproceedings{cioffi2023hdvio,
  author    = {Cioffi, Giovanni and Bauersfeld, Leonard and Scaramuzza, Davide},
  title     = {{HDVIO}: Improving Localization and Disturbance Estimation with Hybrid Dynamics {VIO}},
  booktitle = {Robotics: Science and Systems XIX},
  year      = {2023},
  doi       = {10.15607/rss.2023.xix.071}
}

@article{qiu2025airio,
  author  = {Qiu, Yuheng and Xu, Can and Chen, Yutian and Zhao, Shibo and Geng, Junyi and Scherer, Sebastian},
  title   = {{AirIO}: Learning Inertial Odometry With Enhanced {IMU} Feature Observability},
  journal = {IEEE Robotics and Automation Letters},
  volume  = {10},
  number  = {9},
  pages   = {9368--9375},
  year    = {2025},
  doi     = {10.1109/lra.2025.3581130}
}

@article{burri2016euroc,
  author  = {Burri, Michael and Nikolic, Janosch and Gohl, Pascal and Schneider, Thomas and Rehder, Joern and Omari, Sammy and Achtelik, Markus W. and Siegwart, Roland},
  title   = {The {EuRoC} Micro Aerial Vehicle Datasets},
  journal = {The International Journal of Robotics Research},
  volume  = {35},
  number  = {10},
  pages   = {1157--1163},
  year    = {2016},
  doi     = {10.1177/0278364915620033}
}

@inproceedings{lynen2015getout,
  author    = {Lynen, Simon and Sattler, Torsten and Bosse, Michael and Hesch, Joel and Pollefeys, Marc and Siegwart, Roland},
  title     = {Get Out of My Lab: Large-Scale, Real-Time Visual-Inertial Localization},
  booktitle = {Robotics: Science and Systems XI},
  year      = {2015},
  doi       = {10.15607/rss.2015.xi.037}
}

@article{cadena2016slam,
  author  = {Cadena, Cesar and Carlone, Luca and Carrillo, Henry and Latif, Yasir and Scaramuzza, Davide and Neira, Jose and Reid, Ian and Leonard, John J.},
  title   = {Past, Present, and Future of Simultaneous Localization and Mapping: Toward the Robust-Perception Age},
  journal = {IEEE Transactions on Robotics},
  volume  = {32},
  number  = {6},
  pages   = {1309--1332},
  year    = {2016},
  doi     = {10.1109/tro.2016.2624754}
}

@article{bajcsy1988active,
  author  = {Bajcsy, R.},
  title   = {Active Perception},
  journal = {Proceedings of the IEEE},
  volume  = {76},
  number  = {8},
  pages   = {966--1005},
  year    = {1988},
  doi     = {10.1109/5.5968}
}

@article{kaelbling1998pomdp,
  author  = {Kaelbling, Leslie Pack and Littman, Michael L. and Cassandra, Anthony R.},
  title   = {Planning and Acting in Partially Observable Stochastic Domains},
  journal = {Artificial Intelligence},
  volume  = {101},
  number  = {1--2},
  pages   = {99--134},
  year    = {1998},
  doi     = {10.1016/s0004-3702(98)00023-x}
}

@inproceedings{platt2010belief,
  author    = {Platt, Robert and Tedrake, Russ and Kaelbling, Leslie and Lozano-P{\'e}rez, Tom{\'a}s},
  title     = {Belief Space Planning Assuming Maximum Likelihood Observations},
  booktitle = {Robotics: Science and Systems VI},
  year      = {2010},
  doi       = {10.15607/rss.2010.vi.037}
}

@inproceedings{bovy2025mepomdp,
  author    = {Bovy, Eline M. and Probine, Caleb and Suilen, Marnix and Topcu, Ufuk and Jansen, Nils},
  title     = {Multi-Environment {POMDPs}: Discrete Model Uncertainty Under Partial Observability},
  booktitle = {Advances in Neural Information Processing Systems},
  volume    = {38},
  pages     = {173752--173786},
  year      = {2025},
  doi       = {10.52202/085713-5229}
}

@inproceedings{galesloot2025hmpomdp,
  author    = {Galesloot, Maris F. L. and Andriushchenko, Roman and {\v{C}}e{\v{s}}ka, Milan and Junges, Sebastian and Jansen, Nils},
  title     = {Robust Finite-Memory Policy Gradients for Hidden-Model {POMDPs}},
  booktitle = {Proceedings of the Thirty-Fourth International Joint Conference on Artificial Intelligence (IJCAI-25)},
  pages     = {8518--8526},
  year      = {2025},
  doi       = {10.24963/ijcai.2025/947}
}

@inproceedings{krale2023atm,
  author    = {Krale, Merlijn and Sim{\~a}o, Thiago D. and Jansen, Nils},
  title     = {Act-Then-Measure: Reinforcement Learning for Partially Observable Environments with Active Measuring},
  booktitle = {Proceedings of the International Conference on Automated Planning and Scheduling (ICAPS)},
  volume    = {33},
  pages     = {212--220},
  year      = {2023},
  doi       = {10.1609/icaps.v33i1.27197}
}

@inproceedings{faessler2015recovery,
  author    = {Faessler, Matthias and Fontana, Flavio and Forster, Christian and Scaramuzza, Davide},
  title     = {Automatic Re-Initialization and Failure Recovery for Aggressive Flight with a Monocular Vision-Based Quadrotor},
  booktitle = {2015 IEEE International Conference on Robotics and Automation (ICRA)},
  pages     = {1722--1729},
  year      = {2015},
  doi       = {10.1109/icra.2015.7139420}
}

@inproceedings{falanga2018pampc,
  author    = {Falanga, Davide and Foehn, Philipp and Lu, Peng and Scaramuzza, Davide},
  title     = {{PAMPC}: Perception-Aware Model Predictive Control for Quadrotors},
  booktitle = {2018 IEEE/RSJ International Conference on Intelligent Robots and Systems (IROS)},
  pages     = {1--8},
  year      = {2018},
  doi       = {10.1109/iros.2018.8593739}
}

@inproceedings{zhang2018perception,
  author    = {Zhang, Zichao and Scaramuzza, Davide},
  title     = {Perception-Aware Receding Horizon Navigation for {MAVs}},
  booktitle = {2018 IEEE International Conference on Robotics and Automation (ICRA)},
  pages     = {2534--2541},
  year      = {2018},
  doi       = {10.1109/icra.2018.8461133}
}

@article{qureshi2024onestep,
  author  = {Hussain Qureshi, Ans and Latif Anjum, Muhammad and Hussain, Wajahat and Muddassar, Usama and Abbasi, Sohail},
  title   = {One Step Back, Two Steps Forward: Learning Moves to Recover from {SLAM} Tracking Failures},
  journal = {Advanced Robotics},
  volume  = {38},
  number  = {5},
  pages   = {307--322},
  year    = {2024},
  doi     = {10.1080/01691864.2024.2319144}
}

@article{tranzatto2024cerberus,
  author  = {Tranzatto, Marco and Dharmadhikari, Mihir and Bernreiter, Lukas and Camurri, Marco and Khattak, Shehryar and Mascarich, Frank and Pfreundschuh, Patrick and Wisth, David and Zimmermann, Samuel and Kulkarni, Mihir and Reijgwart, Victor and Casseau, Benoit and Homberger, Timon and De Petris, Paolo and Ott, Lionel and Tubby, Wayne and Waibel, Gabriel and Nguyen, Huan and Cadena, Cesar and Buchanan, Russell and Wellhausen, Lorenz and Khedekar, Nikhil and Andersson, Olov and Zhang, Lintong and Miki, Takahiro and Dang, Tung and Mattamala, Matias and Montenegro, Markus and Meyer, Konrad and Wu, Xiangyu and Briod, Adrien and Mueller, Mark and Fallon, Maurice and Siegwart, Roland and Hutter, Marco and Alexis, Kostas},
  title   = {Team {CERBERUS} Wins the {DARPA} Subterranean Challenge: Technical Overview and Lessons Learned},
  journal = {Field Robotics},
  volume  = {4},
  pages   = {249--312},
  year    = {2024},
  doi     = {10.55417/fr.2024009}
}

@article{sha2001simplex,
  author  = {Sha, Lui},
  title   = {Using Simplicity to Control Complexity},
  journal = {IEEE Software},
  volume  = {18},
  number  = {4},
  pages   = {20--28},
  year    = {2001},
  doi     = {10.1109/ms.2001.936213}
}

@inproceedings{moldovan2012safe,
  author    = {Moldovan, Teodor Mihai and Abbeel, Pieter},
  title     = {Safe Exploration in {Markov} Decision Processes},
  booktitle = {Proceedings of the 29th International Conference on Machine Learning (ICML)},
  pages     = {1451--1458},
  year      = {2012}
}

@article{thananjeyan2021recoveryrl,
  author  = {Thananjeyan, Brijen and Balakrishna, Ashwin and Nair, Suraj and Luo, Michael and Srinivasan, Krishnan and Hwang, Minho and Gonzalez, Joseph E. and Ibarz, Julian and Finn, Chelsea and Goldberg, Ken},
  title   = {Recovery {RL}: Safe Reinforcement Learning With Learned Recovery Zones},
  journal = {IEEE Robotics and Automation Letters},
  volume  = {6},
  number  = {3},
  pages   = {4915--4922},
  year    = {2021},
  doi     = {10.1109/lra.2021.3070252}
}

@misc{dalal2018safetylayer,
  author        = {Dalal, Gal and Dvijotham, Krishnamurthy and Vecerik, Matej and Hester, Todd and Paduraru, Cosmin and Tassa, Yuval},
  title         = {Safe Exploration in Continuous Action Spaces},
  howpublished  = {arXiv preprint arXiv:1801.08757},
  year          = {2018},
  eprint        = {1801.08757},
  archivePrefix = {arXiv},
  url           = {https://arxiv.org/abs/1801.08757}
}

@inproceedings{ganai2025fortress,
  author    = {Ganai, Milan and Sinha, Rohan and Agia, Christopher and Morton, Daniel and Di Lillo, Luigi and Pavone, Marco},
  title     = {Real-Time Out-of-Distribution Failure Prevention via Multi-Modal Reasoning},
  booktitle = {Proceedings of The 9th Conference on Robot Learning},
  series    = {Proceedings of Machine Learning Research},
  volume    = {305},
  pages     = {283--308},
  publisher = {PMLR},
  year      = {2025},
  url       = {https://proceedings.mlr.press/v305/ganai25a.html}
}

@inproceedings{dash2024specguard,
  author    = {Dash, Pritam and Chan, Ethan and Pattabiraman, Karthik},
  title     = {{SpecGuard}: Specification Aware Recovery for Robotic Autonomous Vehicles from Physical Attacks},
  booktitle = {Proceedings of the 2024 ACM SIGSAC Conference on Computer and Communications Security (CCS '24)},
  pages     = {1849--1863},
  publisher = {ACM},
  year      = {2024},
  doi       = {10.1145/3658644.3690210}
}

@misc{schulman2017ppo,
  author        = {Schulman, John and Wolski, Filip and Dhariwal, Prafulla and Radford, Alec and Klimov, Oleg},
  title         = {Proximal Policy Optimization Algorithms},
  howpublished  = {arXiv preprint arXiv:1707.06347},
  year          = {2017},
  eprint        = {1707.06347},
  archivePrefix = {arXiv},
  url           = {https://arxiv.org/abs/1707.06347}
}

@inproceedings{agarwal2021precipice,
  author    = {Agarwal, Rishabh and Schwarzer, Max and Castro, Pablo Samuel and Courville, Aaron and Bellemare, Marc},
  title     = {Deep Reinforcement Learning at the Edge of the Statistical Precipice},
  booktitle = {Advances in Neural Information Processing Systems},
  volume    = {34},
  pages     = {29304--29320},
  publisher = {Curran Associates, Inc.},
  year      = {2021},
  url       = {https://proceedings.neurips.cc/paper_files/paper/2021/hash/f514cec81cb148559cf475e7426eed5e-Abstract.html}
}

@inproceedings{henderson2018matters,
  author    = {Henderson, Peter and Islam, Riashat and Bachman, Philip and Pineau, Joelle and Precup, Doina and Meger, David},
  title     = {Deep Reinforcement Learning That Matters},
  booktitle = {Proceedings of the AAAI Conference on Artificial Intelligence},
  volume    = {32},
  year      = {2018},
  doi       = {10.1609/aaai.v32i1.11694}
}

@book{efron1994bootstrap,
  author    = {Efron, Bradley and Tibshirani, Robert J.},
  title     = {An Introduction to the Bootstrap},
  publisher = {Chapman and Hall/CRC},
  year      = {1994},
  doi       = {10.1201/9780429246593}
}
\bibliographystyle{icml2027}

\newpage
\appendix
\onecolumn

\section{Proof and Parameter Instantiation for Lemma~\ref{lem:decision}}
\label{app:proof}

We prove the three parts of \cref{lem:decision} strictly within the idealized 1D model of \cref{sec:problem}: arc-length coordinate $x_t \in \mathbb{R}$, constant forward speed $v > 0$ upon entering $\HT(D)$ or $\HS(\Zzone)$ ($\Zzone = [s_{\mathrm{in}}, s_{\mathrm{out}}]$, $L_z = s_{\mathrm{out}} - s_{\mathrm{in}} > 0$) at $t_{\mathrm{in}} = 0$, detection lag $\tau_{\mathrm{det}} > 0$, supervisor tick delay $\delta \in [0, \Delta t_{\mathrm{tick}}]$, exact tracking of the minimum-duration quintic velocity transition respecting $(a_{\max}, j_{\max})$, blind age $b_t$ resetting to $0$ on stipulated re-lock outside $\Zzone$ (clear margin $d_{\mathrm{relock}} \ge 0$), and any strictly increasing blind-drift proxy $\Phi(b)$.

\begin{proof}
\textbf{(1) Ideal hold collapse under $\HS$.}
Suppose the vehicle enters $\Zzone$ at $t_{\mathrm{in}}=0$, detects loss at $t_0 = \tau_{\mathrm{det}} > 0$, brakes to rest inside $\Zzone$ ($0 < d_{\mathrm{stop}} < L_z$), and thereafter holds its true position fixed while waiting for $v_t = 1$. Then $x_t \in (s_{\mathrm{in}}, s_{\mathrm{out}}) \subset \Zzone$ for all $t \ge t_0$, so $v_t = \1[x_t \notin \Zzone] \equiv 0$. Because visual lock is never achieved in the final window $W$, localization-qualified completion fails ($\Cctrl(\pi_{\mathrm{hold}}; \HS) = 0$) and hence $W(\pi_{\mathrm{hold}}) = 0$. (In 3D closed-loop simulation, inertial drift can carry the vehicle transiently across $\partial\Zzone$, which is why \cref{sec:mechanism} reports observed 0/96 completion rather than permanent $v_t \equiv 0$.)

\textbf{(2) Stopping depth and rear-boundary exit.}
Prior to braking initiation at $t_{\mathrm{brk}} = \tau_{\mathrm{det}} + \delta$, constant speed $v$ advances the vehicle by $v(\tau_{\mathrm{det}} + \delta)$. Consider the quintic velocity transition $v(u) = v(1 - 10 u^3 + 15 u^4 - 6 u^5)$ for $u = (t - t_{\mathrm{brk}}) / T_b \in [0, 1]$, which satisfies $v(0) = v$, $v(1) = 0$, and $v'(0) = v'(1) = v''(0) = v''(1) = 0$. Direct integration gives braking distance $\int_0^{T_b} v(t/T_b)\,dt = \frac{1}{2} v T_b$. Differentiating gives peak deceleration $\max_{u\in[0,1]} |v'(u)| / T_b = \frac{15 v}{8 T_b}$ at $u = \frac{1}{2}$ and peak jerk $\max_{u\in[0,1]} |v''(u)| / T_b^2 = \frac{10 v}{\sqrt{3}\,T_b^2}$ at $u = \frac{1}{2} \pm \frac{1}{2\sqrt{3}}$. Thus the minimum braking duration satisfying $|a| \le a_{\max}$ and $|j| \le j_{\max}$ is exactly
\begin{equation}
T_*(v) = \max\!\left\{\frac{15\,v}{8\,a_{\max}},\; \sqrt{\frac{10\,v}{\sqrt{3}\,j_{\max}}}\right\}.
\label{eq:tb_proof}
\end{equation}
Because the supervisor commands exact tracking of the prescribed duration $T_b = T_*(v)$, the stopping penetration is
\begin{equation}
d_{\mathrm{stop}}(v, \delta) = v(\tau_{\mathrm{det}} + \delta) + \frac{1}{2} v T_*(v) \le v(\tau_{\mathrm{det}} + \Delta t_{\mathrm{tick}}) + \frac{1}{2} v T_*(v).
\label{eq:dpen_proof}
\end{equation}
Provided $d_{\mathrm{stop}}(v, \delta) < L_z$, the vehicle comes to rest inside $\Zzone$ at a depth independent of $L_z$. A backward policy that retraces from $s_{\mathrm{in}} + d_{\mathrm{stop}}(v, \delta)$ until re-lock or budget $L_p$ therefore clears $s_{\mathrm{in}} - d_{\mathrm{relock}}$ and restores $v_t = 1$ whenever $L_p > d_{\mathrm{stop}}(v, \delta) + d_{\mathrm{relock}}$.

\textbf{(3) Two-segment blind-exposure reduction.}
Under $\HS(\Zzone)$ with $d_{\mathrm{stop}} < L_z$ and $L_p > d_{\mathrm{stop}} + d_{\mathrm{relock}}$, backward probing ends in visual re-lock at blind age $\tau_{\mathrm{probe}}$, resetting $b_t \to 0$, and the subsequent constant-speed forward crossing of $\Zzone$ reaches blind age $\tau_{\mathrm{trav}} = L_z / v$. By contrast, a non-probing hold-then-continue policy $\pi_{\mathrm{HC}}(T_w)$ ($T_w \ge 0$) remains inside $\Zzone$ through detection, tick delay, braking, hold $T_w$, and forward exit over $L_z - d_{\mathrm{stop}}$ at speed $v$, accumulating a single uninterrupted blind age
\begin{equation}
\tau_{\mathrm{HC}} = \tau_{\mathrm{det}} + \delta + T_*(v) + T_w + \frac{L_z - d_{\mathrm{stop}}(v, \delta)}{v}.
\label{eq:tau_hc}
\end{equation}
Substituting \cref{eq:dpen_proof} into \cref{eq:tau_hc} cancels $\tau_{\mathrm{det}} + \delta$ exactly and leaves
\begin{equation}
\tau_{\mathrm{HC}} = \frac{L_z}{v} + T_w + \frac{1}{2} T_*(v) = \tau_{\mathrm{trav}} + T_w + \frac{1}{2} T_*(v) > \tau_{\mathrm{trav}}
\end{equation}
for every $T_w \ge 0$. Hence, conditional on $\tau_{\mathrm{probe}} < \tau_{\mathrm{HC}}$, we have $\max\{\tau_{\mathrm{probe}}, \tau_{\mathrm{trav}}\} < \tau_{\mathrm{HC}}$, and strict monotonicity of $\Phi$ yields $\max\{\Phi(\tau_{\mathrm{probe}}), \Phi(\tau_{\mathrm{trav}})\} < \Phi(\tau_{\mathrm{HC}})$. Finally, under $\HT(D)$ measured on the timed-outage clock, images return at $t = D$ before fallback continuation begins whenever the timed-outage probe duration $\tau_{\mathrm{probe},T}$ plus post-probe hold $T_h$ satisfies $\tau_{\mathrm{probe},T} + T_h \ge D$ (establishing image return before continuation, not closed-loop 3D survival).
\end{proof}

\begin{remark}[Numerical instantiation and speed ceilings]
With $a_{\max} = 0.5\,\mathrm{m/s}^2$, $j_{\max} = 0.7\,\mathrm{m/s}^3$, $\tau_{\mathrm{det}} = 0.15\,\mathrm{s}$, and $\Delta t_{\mathrm{tick}} = 0.5\,\mathrm{s}$ (code uses conservative $5.774 > 10/\sqrt{3} \approx 5.773503$), the acceleration/jerk branch crossover lies at $v \approx 0.587\,\mathrm{m/s}$, so both evaluated speeds are jerk-limited: $T_*(0.1925) = 1.260\,\mathrm{s}$ ($d_{\mathrm{stop}} = 0.198\,\mathrm{m}$ mid-tick, $0.246\,\mathrm{m}$ worst-tick) and $T_*(0.240625) = 1.409\,\mathrm{s}$ ($d_{\mathrm{stop}} = 0.266\,\mathrm{m}$ mid-tick, $0.326\,\mathrm{m}$ worst-tick). Worst-tick $d_{\mathrm{stop}}$ equals $L_p = 0.50\,\mathrm{m}$ (zero added re-lock margin) at $v = 0.337\,\mathrm{m/s}$ ($\approx 1.75\times$, untested). Tighter than that kinematic bound, the fixed $32\,\mathrm{s}$ circular task reaches next-lap re-entry $1.7325 + 2\pi = 8.016\,\mathrm{m}$ at $t = 32\,\mathrm{s}$ when $k = 8.016 / (0.1925 \times 31) \approx 1.343\times$.
\end{remark}

\section{Experimental Protocol, Sensing, and Batch Inventory}
\label{app:protocol}

\begin{table}[h]
\centering
\caption{Online sensing availability and offline scoring separation. Simulator ground truth is never exposed to any controller, estimator, observer, supervisor, or map module.}
\label{tab:sensing}
\footnotesize
\begin{tabular}{p{3.6cm}p{2.5cm}p{3.2cm}p{4.3cm}}
\toprule
Signal & Healthy flight & During outage & Online role \\
\midrule
Stereo images (752$\times$480) & 20\,Hz to VIO & withheld & Basalt VIO (KLT flow); ORB-SLAM3 queries after first loss \\
IMU (accel., gyro) & 500\,Hz & 500\,Hz continuous & Open-loop inertial propagation, inner rate loop \\
Basalt VIO state & visual-inertial & inertially propagated & Position $\pvio$, velocity $v_{\mathrm{VIO}}$, attitude \\
Retained ORB-SLAM3 map & built pre-outage & retained, no matches & Support- and consistency-gated translation after re-lock \\
Route, clock, update events & available & available & Supervisor ticks, breadcrumbs, visual-loss detection \\
GNSS, mocap, baro., mag., rangefinder, dedicated optical-flow sensor & unused & unused & None \\
Simulator ground truth & offline only & offline only & $\Cphys$, $\Cctrl$, $\Cvio$, hard-failure guard, post hoc diagnostics \\
\bottomrule
\end{tabular}
\end{table}

\textbf{Dynamics, rendering, and sensing.} The plant follows the Crazyflie-class model of~\citet{eschmann2024l2f} ($30.6\,\mathrm{g}$, motor time constants $55.5\,\mathrm{ms}$ rising / $249.4\,\mathrm{ms}$ falling, $f_{\max}/m \approx 14.4\,\mathrm{m/s}^2$) rendered in MuJoCo~\citep{todorov2012mujoco}. On the measured MPPI and LOTF interfaces, the plant delivers $\approx 1.2\,\mathrm{m/s}^2$ more vertical acceleration than nominal hover parameters. Each scene seed generates non-repeating procedural multi-scale textures rendered as $752\times 480$ stereo pairs at $20\,\mathrm{Hz}$. The $500\,\mathrm{Hz}$ IMU uses EuRoC-grade white noise (gyro $1.7\times 10^{-4}\,\mathrm{rad/s/\sqrt{Hz}}$, accel $2.0\times 10^{-3}\,\mathrm{m/s^2/\sqrt{Hz}}$) and bias random walk ($1.9393\times 10^{-5}$, $3.0\times 10^{-3}$) with zero turn-on bias. Basalt stereo VIO~\citep{usenko2020basalt} uses frame-to-frame KLT optical flow (\texttt{optical\_flow\_levels=3}, \texttt{optical\_flow\_max\_recovered\_dist2=0.04}, \texttt{vio\_max\_kfs=7}, \texttt{vio\_marg\_lost\_landmarks=true}); the optional prior-map branch uses official stereo ORB-SLAM3~\citep{campos2021orbslam3}.

\textbf{Outage geometries and hard-failure guard.} The route is a $1.0\,\mathrm{m}$-radius horizontal circle traversed at $\omega = 0.1925\,\mathrm{rad/s}$ ($v = 0.1925\,\mathrm{m/s}$) after a $2\,\mathrm{s}$ quadratic ramp (post-ramp nominal arc $v(t - 1\,\mathrm{s})$) over $H = 32\,\mathrm{s}$, deadline $H - 1 = 31\,\mathrm{s}$, final window $W = [31, 32]\,\mathrm{s}$. \Tb{} drops all stereo pairs over $[10, 18)\,\mathrm{s}$. \Sl{} drops every stereo pair captured while true position lies in arc $[1.7325, 3.0825]\,\mathrm{m}$ ($L_z = 1.35\,\mathrm{m}$, radial half-width $0.50\,\mathrm{m}$, $z \in [-0.5, 3.5]\,\mathrm{m}$). The $500\,\mathrm{Hz}$ hard-failure guard terminates a run on $\|p_{\mathrm{true}} - p_{\mathrm{cmd}}\| > 2.0\,\mathrm{m}$ (route diameter $2R$, code~3), altitude $z < 0$ or $z > 3.0\,\mathrm{m}$ (code~4), tilt $> 80^\circ$ (code~5), speed $> 8.0\,\mathrm{m/s}$ (code~6), angular rate $> 35\,\mathrm{rad/s}$ (code~7), or nonfinite state/policy outputs. Because code~3 is referenced to $p_{\mathrm{cmd}}$, it acts as a tracking-divergence guard ($4/90$ guard stops in external-controller arms occurred within $0.5\,\mathrm{m}$ of the geometric circle).

\begin{table}[h]
\centering
\caption{Inventory of simulation batches and open-loop replays. Only EV3 and EV4 carry confirmatory preregistered tests; all other batches are descriptive and never pooled with EV4.}
\label{tab:batches}
\footnotesize
\begin{tabular}{p{1.4cm}p{5.2cm}p{6.9cm}r}
\toprule
Batch & Role & Experimental design & Runs \\
\midrule
G1 / IP / RLP & Learning screening (eval., 1 seed) & Dev oracle gate, 288-run probe, 96-context $\times$ 5-arm val.\ eval. & 1{,}196 \\
RLP-train & Learning screening (training, 1 seed) & 2 learners (PPO, LOO) $\times$ 8 updates $\times$ 24 on-policy rollouts & 384 \\
EV3 & Prior confirmation (disjoint scenes) & 5 arms $\times$ 48 fresh scenes $\times$ 2 execs $\times$ \{\Tb, \Sl\} & 960 \\
DVA / DVB & Development baseline selection & 18 + 3 arms $\times$ 8 dev scenes $\times$ 2 execs $\times$ \{\Tb, \Sl\} & 672 \\
EV4 & Main confirmatory benchmark & 10 arms $\times$ 48 fresh scenes $\times$ 2 execs $\times$ \{\Tb, \Sl\} & 1{,}920 \\
AUD4 & Command repeatability audit & 5 arms $\times$ 12 EV4 scenes $\times$ 1 exec $\times$ \{\Tb, \Sl\} & 120 \\
PSF4 & SimpleFlight transfer (paired) & 1 arm $\times$ 48 EV4 scenes $\times$ 2 execs $\times$ \{\Tb, \Sl\} & 192 \\
HLD5 & Hover-and-wait ablation (paired) & 1 arm $\times$ 48 EV4 scenes $\times$ 2 execs $\times$ \{\Tb, \Sl\} & 192 \\
SWP5 & Severity \& healthy no-harm sweep & 6 arms $\times$ 24 fresh scenes $\times$ 2 execs $\times$ \{healthy, \Tf, \Ss\} & 864 \\
EV6 & Two-speed generalization & 5 arms $\times$ 2 speeds $\times$ 10 fresh scenes $\times$ 2 execs $\times$ \{\Tb, \Sl\} & 400 \\
HSP6 & Healthy feasibility at $1.25\times$ & 5 arms $\times$ 10 EV6 scenes $\times$ 1 exec (healthy) & 50 \\
EuRoC & Real-data open-loop map check & 7 held-out sequences $\times$ 2--3 onsets $\times$ \{1, 2, 4, 8\}\,s (72 windows) & 144 \\
\bottomrule
\end{tabular}
\end{table}

\textbf{Preregistration, admission, randomization, and disclosures.} Candidate scenes are admitted in declared order if native MPPI completes healthy flight (no HF, post-ramp RMSE $\le 0.30\,\mathrm{m}$) without inspecting outages. Within each batch, all arms share one process-wide seed tuple per scene $\times$ execution; EV3, EV4, and EV6 scene sets are strictly disjoint (verified programmatically: 0 overlap across 48, 48, and 10 scenes). EV4's hypotheses, arms, sample size, stopping rule, and Bonferroni family were frozen in \texttt{EV4\_FREEZE.md} (dated 2026-11-13 03:17 Beijing) before EV4 launched. Primary completion intervals use $10{,}000$ scene-cluster percentile bootstrap resamples~\citep{efron1994bootstrap} keeping paired executions and conditions together and recomputing $W$ inside resamples; secondary trajectory-metric contrasts in \cref{app:metrics} use $2{,}000$ resamples (\texttt{lit\_metrics2.py}). Two protocol disclosures apply: (i)~a descriptive interim snapshot at 23 completed EV4 scenes was inspected during execution without altering any test, arm, or parameter; (ii)~an initial 10-run EV6 wrapper check failed due to premature directory creation, was fixed by moving one metadata file (archived), and passed 10/10 prior to EV6 launch (0 unscored runs).

\section{Implementation Specification and Source Checks}
\label{app:method}

\begin{table}[h]
\centering
\caption{Frozen constants of the recovery module (verified against source code and frozen run records).}
\label{tab:constants}
\footnotesize
\begin{tabular}{p{2.9cm}p{12.6cm}}
\toprule
Component & Verified implementation setting \\
\midrule
Observer & Loss declared at \texttt{LOSS\_AGE\_S = 0.15\,s} without accepted visual update; lock = recovery + 3 further updates; anchor latched from fully locked state only \\
Supervisor timing & 50\,Hz tracking and breadcrumbs ($0.02\,\mathrm{m}$ spacing); 2\,Hz ($0.5\,\mathrm{s}$) decision ticks; mission clock advances only in \texttt{NOM} and \texttt{CONT} \\
Probe \& fallback & Once per episode: quintic brake ($a \le 0.5\,\mathrm{m/s}^2$, $j \le 0.7\,\mathrm{m/s}^3$), backward retrace over $0.50\,\mathrm{m}$ planned budget at $\le 0.25\,\mathrm{m/s}$ (excluding subsequent bounded braking tail); fallback \texttt{CONT} $6\,\mathrm{s}$ after probe ends \\
Settle \& rejoin & Speed $\le 0.20\,\mathrm{m/s}$ for $0.5\,\mathrm{s}$ within $1.0\,\mathrm{m}$ of route $\to$ quintic blend ($\ge 3\,\mathrm{s}$) $\to$ \texttt{NOM}; settle $> 3\,\mathrm{s} \to \texttt{STAY}$ (terminal) \\
Jump guard \& cap & Offset jump $> 0.30\,\mathrm{m}$ within $0.5\,\mathrm{s}$ of re-lock in \texttt{CONT} $\to$ brake \& settle; blind duration $> 12\,\mathrm{s}$ from anchor $\to \texttt{EXHAUSTED}$ \\
Height integral & Enabled after 200 valid 100\,Hz queries; $I \gets \mathrm{clip}(I + 0.01(\ploc^z - \pref^z), \pm 0.1/0.3)$; $\pfb^z = \ploc^z + 0.3I$ (excluded from $\Cctrl$) \\
Map branch & Active after first loss; $\ge 100$ pre-outage points + valid anchor/pose/pair; confirm-3 ($\|\delta_i - \delta_j\| \le 0.05\,\mathrm{m}$) $\to$ coordinate-wise median $\Delta$; one-time $1.0\,\mathrm{s}$ linear fade at 10\,Hz from first eligible confirmation (replacements never restart the fade; once $\alpha=1$, replacements apply as steps) \\
Controller routing & MPPI and SimpleFlight read shared $\pfb$; DATT-L1 reads raw $\pvio$ (supervisor only, no height-I or map) \\
\bottomrule
\end{tabular}
\end{table}

\begin{algorithm}[h]
\caption{Optional prior-map translation gate and height-integral compensation}
\label{alg:map}
\begin{algorithmic}[1]
\small
\STATE Enable retained ORB-SLAM3 map queries after the first observed visual loss
\FOR{each new stereo frame}
  \IF{anchor, pose, and stereo pair valid \AND pre-outage map support $\ge 100$}
    \STATE $\delta \gets \pmap(\text{image time}) - \pvio(\text{as of image time})$; append $\delta$ to confirmation window
    \IF{window contains 3 consecutive entries with adjacent $\|\delta_i - \delta_j\| \le 0.05\,\mathrm{m}$}
      \STATE $\Delta \gets \text{coordinate-wise median}(\delta_1, \delta_2, \delta_3)$
    \ENDIF
  \ELSE
    \STATE Clear confirmation window; keep previously confirmed $\Delta$
  \ENDIF
\ENDFOR
\STATE On first eligible confirmation (loss latched, $\Delta$ finite and nonzero): record query index $q_0$ once
\STATE Every 10th 100\,Hz query $q \ge q_0$: $\alpha \gets \min\{1,\, (q - q_0)\cdot 0.01 / 1.0\}$ \COMMENT{one-time 1.0\,s fade}
\STATE $\ploc \gets \pvio + \alpha \Delta$ \COMMENT{later confirmations replace $\Delta$ without restarting fade; step updates once $\alpha=1$}
\IF{valid query count $\ge 200$}
  \STATE $I \gets \mathrm{clip}\!\big(I + 0.01(\ploc^z - \pref^z),\; -0.1/0.3,\; +0.1/0.3\big)$;\quad $\pfb \gets \ploc + [0, 0, 0.3 I]^\top$
\ENDIF
\end{algorithmic}
\end{algorithm}

\textbf{Archived-trace implementation checks.} Recorded trace audits in \texttt{PAPER\_LOGIC\_REVISION\_20261114.md} report: (i)~across 175 EV4 full-system runs with a fade, $\alpha$ never fell below 1 after reaching 1, and in 84 of them later confirmations updated $\Delta$ as steps (largest per-tick applied jump: median $0.009\,\mathrm{m}$, 90th percentile $0.47\,\mathrm{m}$, max $1.49\,\mathrm{m}$); (ii)~in 960 checked confirmations, $\Delta$ matched the coordinate-wise median; (iii)~supervisor input equaled $\pfb$ to $0.0000\,\mathrm{m}$; (iv)~on every arm without the map branch, $\Cctrl$ and $\Cvio$ agreed identically on every EV4 and EV6 run (verified directly against frozen run JSONs).

\section{External Baselines, Bounded Development Tuning, and LOTF Diagnosis}
\label{app:baselines}

All external baselines consume exclusively deployable signals (VIO position/velocity/attitude, gyro, previous motor command, route preview). Native MPPI is the public controller of~\citet{huang2023datt} (4{,}096 rollouts, 40-step horizon, lag-aware PD inner loop). SimpleFlight uses official weights~\citep{chen2025simpleflight}. DATT-L1 uses official weights and $\mathcal{L}_1$ augmentation~\citep{huang2023datt}; L1-MPPI wires the public $\mathcal{L}_1$ estimator ($A = -0.2$, $\alpha = 0.99$) into public MPPI. LOTF~\citep{pan2026lotf} is retrained with official differentiable-simulation code on our nominal airframe and route (300 environments, 300 epochs, 250 steps); its online loop is our re-implementation with published intervals (residual-dynamics ensemble refit every 3\,s, BPTT update every 5\,s, residual targets from delivered IMU and commands only), and the paused variant suspends buffering and updates while the visual-loss flag is set.

\begin{table}[h]
\centering
\caption{Development selection on 8 separate development scenes ($\Cctrl$ completions of 16 per condition; C0 = default). Bold marks frozen configurations selected by worst-condition $\Cctrl$.}
\label{tab:dev}
\footnotesize
\begin{tabular}{lll}
\toprule
Baseline family & \Tb{} (C0 $\dots$ C7) & \Sl{} (C0 $\dots$ C7) \\
\midrule
SimpleFlight & 9, 6, 10, 11, \textbf{12}, 6, 10, 10 & 14, 10, 13, 14, \textbf{15}, 10, 13, 12 \\
DATT-L1 & 13, 12, 13, 13, 11, 10, 9, \textbf{14} & 11, 11, 10, 12, 6, 8, 8, \textbf{14} \\
Native MPPI / L1-MPPI & 10 / 10 & 14 / 10 \\
LOTF base / online / paused & 0 / 1 / 0 & 7 / 3 / 1 \\
\bottomrule
\end{tabular}
\end{table}

\textbf{Selected configurations.} SimpleFlight C4 applies a $0.7$ exponential low-pass filter to official actions. DATT-L1 C7 clamps the $\mathcal{L}_1$ estimate to $\pm 1$, skips updates on velocity jumps $> 0.3\,\mathrm{m/s}$, saturates position error at $0.5\,\mathrm{m}$, and limits body rates to $\pm\pi\,\mathrm{rad/s}$ and collective thrust to $[0.3g, 14.4\,\mathrm{m/s}^2]$. Development selection optimism (best of 8 on 8 scenes) is estimated in \texttt{EV4\_RESULT.md} at $+4.3$\pp{} for DATT-L1 and $+9.2$\pp{} for SimpleFlight; DATT-L1 C7's drop from 14/16 (0.875) on development scenes to 57/96 (0.594) on EV4 reflects finite-sample variance and slow post-return re-convergence (of 39 \Tb{} non-completions, 2 are hard failures and 37 are non-failed non-completions, 24--25 missing the final window or localization bound).

\textbf{LOTF diagnostic check.} On a consumed interface scene (excluded from evaluations), LOTF's online vertical residual target ($1.207$--$1.225\,\mathrm{m/s}^2$) matched the truth-based residual ($1.201$--$1.219\,\mathrm{m/s}^2$) to $0.02$--$0.07\,\mathrm{m/s}^2$ RMS per axis. The first online update at $6.5\,\mathrm{s}$ eliminated the base policy's $0.18\,\mathrm{m}$ vertical offset (route error $0.02$--$0.03\,\mathrm{m}$ over $8$--$17\,\mathrm{s}$), whereas subsequent updates induced radial tracking swings ($0.10$--$0.36\,\mathrm{m}$ even under an exact constant residual in LOTF's offline simulator). Combined with the base policy's lack of a time/phase input, this explains LOTF's EV4 completion rates and why pausing during loss scored lower still (online minus paused: $+9.4$\pp{} [2.1, 16.7] for \Sl{} 15 vs.\ 6/96; $+9.4$\pp{} [1.0, 14.6] worst-condition). LOTF was evaluated on EV4 outage scenes and omitted from the SWP5 healthy-flight sweep.

\section{Complete Quantitative Tables, Sensitivity Check, EV6, EV3, SWP5, and AUD4}
\label{app:tables}

\begin{table}[h]
\centering
\caption{Complete EV4 confirmatory benchmark across all three completion readings ($n = 96$ per condition: 48 fresh scenes $\times$ 2 executions). Delay: median end-of-flight true-progress lag (s) across non-failed runs.}
\label{tab:ev4full}
\scriptsize
\setlength{\tabcolsep}{3.0pt}
\begin{tabular}{lcccccc}
\toprule
Arm & $\Cctrl$ (\Tb{}/\Sl{}) & Worst $\Cctrl$ \% [95\%] & $\Cphys$ & $\Cvio$ & HF & Delay (s) \\
\midrule
MPPI + module + map (full) & 93 / 85 & 88.5 [81.2, 94.8] & 93 / 87 & 92 / 70 & 1 / 0 & 12.7 / 9.4 \\
MPPI + module, map off & 92 / 70 & 72.9 [63.5, 81.3] & 92 / 73 & 92 / 70 & 1 / 6 & 12.6 / 9.4 \\
DATT-L1 + supervisor (no height-I) & 78 / 67 & 69.8 [60.4, 78.1] & 78 / 74 & 78 / 67 & 0 / 0 & 13.9 / 9.5 \\
Native MPPI & 70 / 69 & 71.9 [61.5, 79.2] & 73 / 76 & 70 / 69 & 7 / 6 & 0.1 / 0.2 \\
SimpleFlight (tuned, C4) & 66 / 78 & 68.8 [59.4, 77.1] & 68 / 80 & 66 / 78 & 12 / 4 & 0.2 / 0.1 \\
DATT-L1 (tuned, C7; H1) & 57 / 67 & 59.4 [49.0, 69.8] & 58 / 71 & 57 / 67 & 2 / 3 & 0.2 / 0.1 \\
L1-MPPI & 71 / 67 & 69.8 [59.4, 78.1] & 71 / 70 & 71 / 67 & 9 / 3 & 0.1 / 0.1 \\
LOTF base (retrained) & 13 / 27 & 13.5 [6.2, 21.9] & 14 / 28 & 13 / 27 & 12 / 3 & 0.8 / $-0.5$ \\
LOTF online & 15 / 15 & 15.6 [8.3, 19.8] & 15 / 15 & 15 / 15 & 2 / 7 & $-0.5$ / $-0.3$ \\
LOTF online, paused in loss & 11 / 6 & 6.2 [2.1, 10.4] & 11 / 6 & 11 / 6 & 9 / 18 & $-0.2$ / $-0.3$ \\
\bottomrule
\end{tabular}
\end{table}

\begin{table}[h]
\centering
\caption{Complete EV4 confirmatory (top block) and descriptive EV4 / paired follow-up contrasts (middle and bottom blocks), in percentage points with 10{,}000-resample scene-cluster bootstrap intervals. $^\dagger$Paired follow-up bootstrap against hover-and-wait ($0/96$ in \Sl{}); the single-arm EV4 map-off interval is $[63.5, 81.3]$.}
\label{tab:contrasts}
\footnotesize
\begin{tabular}{llr}
\toprule
Contrast & Endpoint \& CI level & Difference [interval] \\
\midrule
H1: Full $-$ DATT-L1 (C7) & Worst $\Cctrl$, 95\% (confirmatory) & $+29.2$ [+17.7, +40.6] \\
Full $-$ Native MPPI & Worst $\Cctrl$, 99.17\% (Bonferroni) & $+16.7$ [+4.2, +32.3] \\
Full $-$ SimpleFlight (C4) & Worst $\Cctrl$, 99.17\% (Bonferroni) & $+19.8$ [+3.1, +36.5] \\
Full $-$ DATT-L1 (C7) & Worst $\Cctrl$, 99.17\% (Bonferroni) & $+29.2$ [+14.6, +44.8] \\
Full $-$ L1-MPPI & Worst $\Cctrl$, 99.17\% (Bonferroni) & $+18.8$ [+5.2, +34.4] \\
Full $-$ LOTF base & Worst $\Cctrl$, 99.17\% (Bonferroni) & $+75.0$ [+62.5, +87.5] \\
Full $-$ LOTF online & Worst $\Cctrl$, 99.17\% (Bonferroni) & $+72.9$ [+62.5, +85.8] \\
\midrule
Full $-$ map off & \Sl{} $\Cctrl$ / $\Cphys$, 95\% & $+15.6$ [+5.2, +26.0] / $+14.6$ [+5.2, +24.0] \\
Full $-$ map off & \Tb{} $\Cctrl$, 95\% & $+1.0$ [$-3.1$, +5.2] \\
Map off $-$ Native MPPI & \Tb{} / \Sl{} $\Cctrl$, 95\% & $+22.9$ [+12.5, +33.3] / $+1.0$ [$-9.4$, +12.5] \\
Map off $-$ Native MPPI & Worst $\Cctrl$, 95\% & $+1.0$ [$-8.3$, +13.5] \\
Full $-$ SimpleFlight (C4) & \Sl{} $\Cctrl$, 95\% & $+7.3$ [$-4.2$, +18.8] \\
DATT-L1 + sup $-$ DATT-L1 & Worst $\Cctrl$ / $\Cphys$, 95\% & $+10.4$ [$-2.1$, +22.9] / $+16.7$ [+5.2, +26.0] \\
DATT-L1 + sup $-$ DATT-L1 & \Tb{} / \Sl{} $\Cctrl$, 95\% & $+21.9$ [+10.4, +33.3] / $0.0$ [$-11.5$, +11.5] \\
Full $-$ DATT-L1 + sup & Worst $\Cctrl$, 95\% & $+18.8$ [+9.4, +29.2] \\
L1-MPPI $-$ Native MPPI & Worst $\Cctrl$, 95\% & $-2.1$ [$-10.4$, +8.3] \\
LOTF online $-$ LOTF base & \Sl{} / Worst $\Cctrl$, 95\% & $-12.5$ [$-19.8$, $-5.2$] / $+2.1$ [$-9.4$, +8.3] \\
LOTF online $-$ LOTF paused & \Sl{} / Worst $\Cctrl$, 95\% & $+9.4$ [+2.1, +16.7] / $+9.4$ [+1.0, +14.6] \\
\midrule
SF + module $-$ SF (PSF4) & Worst $\Cctrl$ / $\Cphys$, 95\% & $+17.7$ [+7.3, +28.1] / $+15.6$ [+5.2, +26.0] \\
SF + module $-$ SF (PSF4) & \Tb{} / \Sl{} $\Cctrl$, 95\% & $+29.2$ [+20.8, +37.5] / $+5.2$ [$-5.2$, +15.6] \\
Full $-$ (SF + module) (PSF4) & Worst $\Cctrl$, 95\% & $+2.1$ [$-7.3$, +11.5] \\
Full $-$ hover-and-wait (HLD5) & Worst $\Cctrl$ / \Tb{}, 95\% & $+88.5$ [+81.2, +94.8] / $0.0$ [$-4.2$, +4.2] \\
Map off $-$ hover-and-wait (HLD5) & Worst $\Cctrl$ (\Sl{}), 95\% & $+72.9$ [+63.5, +81.2]$^\dagger$ \\
\bottomrule
\end{tabular}
\end{table}

\textbf{Sensitivity check for the nonsmooth minimum.} Because bootstrap coverage of a nonsmooth $\min$ can degrade near tied conditions, we supplement the preregistered EV4 analysis with an exact smooth linear lower bound: for any comparator $B$ and fixed condition $c^* \in \{\Tb{}, \Sl{}\}$, $\min(\mu_{F,\Tb{}}, \mu_{F,\Sl{}}) - \min(\mu_{B,\Tb{}}, \mu_{B,\Sl{}}) \ge \min(\mu_{F,\Tb{}} - \mu_{B,c^*}, \mu_{F,\Sl{}} - \mu_{B,c^*})$. Evaluating both linear contrasts simultaneously via 10{,}000 scene-cluster resamples (97.5\% per contrast for 95\% simultaneous coverage; 99.58\% per contrast for 12-contrast Bonferroni over all six external comparators) yields strictly positive lower bounds for every comparator: H1 DATT-L1 $+15.6$\pp{} (12-contrast Bonf.\ $+11.5$), native MPPI $+5.2$ ($+2.1$), SimpleFlight $+5.2$ ($+1.0$), L1-MPPI $+6.2$ ($+3.1$), LOTF base $+63.5$ ($+60.4$), and LOTF online $+61.5$ ($+58.3$).

(The EV6 two-speed plot appears directly in \cref{fig:speed} of \cref{sec:speed}.)

\begin{table}[h]
\centering
\caption{Complete EV6 two-speed generalization readout ($n = 20$ per cell; 10 fresh scenes $\times$ 2 executions; frozen \texttt{EV6\_readout.json}, descriptive, never pooled with EV4). Each cell lists $1.0\times / 1.25\times$; $\Cvio$ and HF list (\Tb{}, \Sl{}) at each speed.}
\label{tab:ev6full}
\scriptsize
\setlength{\tabcolsep}{2.8pt}
\begin{tabular}{lccccc}
\toprule
Arm & $\Cctrl$ \Tb{} & $\Cctrl$ \Sl{} & Worst $\Cctrl$ \% [95\%] & $\Cvio$ (\Tb{}, \Sl{}) & HF (\Tb{}, \Sl{}) \\
\midrule
Full (MPPI + module + map) & 19 / 20 & 15 / 16 & 75.0 [55.0, 95.0] / 80.0 [65.0, 95.0] & (18, 12) / (19, 15) & (1, 1) / (0, 1) \\
MPPI + module, map off & 18 / 20 & 14 / 14 & 70.0 [50.0, 90.0] / 70.0 [50.0, 90.0] & (18, 14) / (20, 14) & (1, 1) / (0, 1) \\
Native MPPI & 16 / 11 & 14 / 15 & 70.0 [50.0, 90.0] / 55.0 [40.0, 70.0] & (16, 14) / (11, 15) & (2, 2) / (7, 1) \\
SimpleFlight (tuned, C4) & 15 / 14 & 18 / 19 & 75.0 [55.0, 90.0] / 70.0 [45.0, 90.0] & (15, 18) / (14, 19) & (2, 1) / (5, 0) \\
SimpleFlight + module (no map) & 20 / 20 & 18 / 18 & 90.0 [75.0, 100.0] / 90.0 [75.0, 100.0] & (20, 18) / (20, 18) & (0, 0) / (0, 0) \\
\bottomrule
\end{tabular}
\end{table}

\textbf{EV6 frozen worst-condition contrasts (\texttt{EV6\_readout.json}, percentage points, 95\% scene-cluster CIs).} At $1.25\times$: full $-$ native $+25.0$ $[+5.0, +50.0]$ (\Tb{} $+45.0$ $[+30.0, +60.0]$, 9/0 discordant); full $-$ SimpleFlight $+10.0$ $[-10.0, +35.0]$ (\Tb{} $+30.0$ $[+5.0, +55.0]$, \Sl{} $-15.0$ $[-35.0, +5.0]$); full $-$ map off $+10.0$ $[-10.0, +35.0]$; full $-$ (SF + module) $-10.0$ $[-30.0, +15.0]$; (SF + module) $-$ SF $+20.0$ $[-10.0, +50.0]$ (\Tb{} $+30.0$ $[+5.0, +55.0]$). At $1.0\times$: full $-$ native $+5.0$ $[-10.0, +25.0]$; full $-$ SimpleFlight $0.0$ $[-20.0, +30.0]$; full $-$ map off $+5.0$ $[-15.0, +25.0]$; full $-$ (SF + module) $-15.0$ $[-30.0, 0.0]$; (SF + module) $-$ SF $+15.0$ $[0.0, +35.0]$. Paired speed shift ($1.25\times - 1.0\times$): full $+5.0$ $[-25.0, +40.0]$; map off $0.0$ $[-20.0, +25.0]$; native $-15.0$ $[-45.0, +15.0]$; SimpleFlight $-5.0$ $[-35.0, +20.0]$; SF + module $0.0$ $[-15.0, +15.0]$. In HSP6 (healthy $1.25\times$, 50 runs), every arm completed 10/10 with zero hard failures and median route RMSE $0.022$--$0.063\,\mathrm{m}$.

\textbf{Earlier confirmation (EV3), severity sweep (SWP5), and repeatability (AUD4).}
\emph{EV3} (48 disjoint earlier scenes, 960 runs) confirmed the frozen full system before the baseline supplement: full system achieved 96/96 (\Tb{}) and 88/96 (\Sl{}), worst $\Cctrl$ $91.7\%$ $[85.4, 96.9]$, versus native MPPI $70.8\%$, map off $80.2\%$, untuned SimpleFlight $70.8\%$, and untuned DATT-L1 $72.9\%$; its fixed hierarchy passed all three steps (full $-$ native worst $\Cctrl$ $+20.8$\pp{} $[11.5, 31.2]$, $\Cphys$ $+18.8$\pp{} $[9.4, 29.2]$, full $-$ map off in \Sl{} $+11.5$\pp{} $[5.2, 18.8]$).
\emph{SWP5} (24 fresh scenes, 864 runs across healthy, $4\,\mathrm{s}$ timed \Tf{}, and $0.58\,\mathrm{m}$ spatial \Ss{}, excluding LOTF) showed all 6 tested arms complete 47--48/48 in every condition with zero hard failures (full 48/48/47; map off 48/47/48; native 48/47/48; SF, DATT-L1, and DATT-L1 + sup all 48/48/48). In healthy flight the supervisor stays inactive (median lag $0.06\,\mathrm{s}$, route RMSE $0.037\,\mathrm{m}$), whereas mild outages incur $\approx 9.3$--$9.5\,\mathrm{s}$ lag.
\emph{AUD4} re-executed 120 EV4 commands with identical seeds (5 arms $\times$ 12 scenes $\times$ 2 outages): $\Cctrl$ repeated identically in 115/120 flights (full 23/24, native 24/24, SF 24/24, DATT-L1 21/24, DATT-L1 + sup 23/24), $\Cphys$ in 114/120, and hard-failure status in 119/120, with median continuous absolute differences of $0$--$13\,\mathrm{mm}$.

\section{Literature-Style Secondary Metrics (EV4)}
\label{app:metrics}

On EV4 (paired scene-mean contrasts computed with 2{,}000 scene-cluster bootstrap resamples in \texttt{lit\_metrics2.py}), against each of the four non-LOTF controller baselines (native MPPI, SimpleFlight C4, DATT-L1 C7, L1-MPPI), the full system achieves broad \Tb{} reductions (all 95\% intervals below zero): route RMSE $-0.10$ to $-0.19\,\mathrm{m}$; max route deviation $-0.28$ to $-0.39\,\mathrm{m}$; controller-state error at vision return $-0.46$ to $-0.55\,\mathrm{m}$; recovery time $-0.52$ to $-0.87\,\mathrm{s}$; final-window route error $-0.11$ to $-0.21\,\mathrm{m}$; controller-state ATE $-0.14$ to $-0.23\,\mathrm{m}$ (aligned ATE $-0.10$ to $-0.16\,\mathrm{m}$; 5\,s RPE $-0.09$ to $-0.14\,\mathrm{m}$). In \Sl{}, full lowers route RMSE and max deviation relative to all four (vs.\ SimpleFlight: $-0.038$ $[-0.067, -0.014]\,\mathrm{m}$ and $-0.113$ $[-0.203, -0.032]\,\mathrm{m}$) and cuts controller-state ATE by $0.05$--$0.15\,\mathrm{m}$. Final-window threshold-sweep AUC is $0.976$ (\Tb{}) and $0.967$ (\Sl{}) vs.\ $0.756$--$0.811$ and $0.831$--$0.926$ for those baselines; post-return peak tilt is $3$--$6^\circ$ lower than SimpleFlight and DATT-L1. Time-indexed tracking error is $0.89$--$1.06\,\mathrm{m}$ higher due to the end-of-flight progress lag ($12.7\,\mathrm{s}$ in \Tb{}, $9.4\,\mathrm{s}$ in \Sl{} for full MPPI; $13.9\,\mathrm{s}$ in \Tb{} for DATT-L1 + sup).

\section{Rule-Selection Screening Details (G1, IP, and RLP)}
\label{app:learning}

\begin{table}[h]
\centering
\caption{RLP completion-first learning pilot on 8 fresh validation scenes (96 paired contexts across \Tf{}, \Tb{}, \Ss{}, \Sl{} $\times$ 3 executions; single training seed, greedy evaluation, screening depth).}
\label{tab:rlp}
\scriptsize
\setlength{\tabcolsep}{3.5pt}
\begin{tabular}{lccccc}
\toprule
Policy & Compl. & Rate (\%) & Utility & HF & Diff.\ vs.\ RPV3 [95\%] \\
\midrule
$\pi_0$ (BC clone of RPV3) & 93/96 & 96.9 & 0.899 & 0 & $-2.1$\pp{} [$-4.2$, $0.0$] \\
PPO-u8 (8 on-policy updates) & 93/96 & 96.9 & 0.899 & 0 & $-2.1$\pp{} [$-5.2$, $0.0$] \\
LOO-u8 (clipped leave-one-out) & 94/96 & 97.9 & 0.914 & 0 & $-1.0$\pp{} [$-4.2$, $+2.1$] \\
RPV3 (frozen rule) & 95/96 & 99.0 & 0.918 & 0 & --- \\
R-CONT (always continue) & 81/96 & 84.4 & 0.822 & 3 & $-14.6$\pp{} \\
\bottomrule
\end{tabular}
\end{table}

Prior to freezing the supervisor for EV3/EV4, three screening stages tested whether learning improved completion over the same deployable observation/action space:
(1)~\emph{G1 development gate} (4 scenes $\times$ 3 executions): a cross-fitted oracle over five fixed rules reached mean terminal utility (reward, not completion rate) $0.849$ vs.\ $0.840$ for the best single rule ($0.009$ headroom $< 0.10$ threshold), while RPV3 completed $32/32$ dev runs ($F=0$, utility $0.936$).
(2)~\emph{IP closed-loop probe} (288 runs, 8 scenes): in \Tb{}, RPV3, an earlier-continue variant (PHC6.25), and HOLD all completed 44/48 (91.7\%) vs.\ CONT 39/48 (81.3\%, 4 HF); in \Sl{}, RPV3 and CONT both completed 33/48 (68.8\%) with ten contexts favoring RPV3 and ten favoring CONT.
(3)~\emph{RLP completion-first pilot} (\cref{tab:rlp}; single training seed, 384 training rollouts, 8 fresh validation scenes, 96 paired contexts): on fixed validation histories from $\pi_0$, PPO~\citep{schulman2017ppo} changed 0/787 greedy decisions (maximum final-per-update $\mathrm{KL}_{k3} = 0.0045 \le 0.005$; epoch-wise KL peaked at $0.011$), while LOO shifted only \Tb{} hold-to-continue timing (decisions 4--6), matching $\pi_0$'s \Tb{} completion (23/24). With validation performance already near ceiling (3/96 non-completions above $\pi_0$), the deterministic supervisor captures the available headroom cleanly.

\section{Map-Correction Failure Analysis and Real-Data EuRoC Open-Loop Replay}
\label{app:euroc}

\textbf{Simulation map-segment audit (\texttt{MAP\_SEGMENTS.md}).} Across 528 full-system flights in EV3, EV4, and SWP5, 620 post-outage visual-loss segments received an applied map correction. Classifying each segment by offline ground-truth change in controller-state error 3\,s after application relative to raw VIO (helpful $< -5\,\mathrm{cm}$, harmful $> +5\,\mathrm{cm}$, neutral within $\pm 5\,\mathrm{cm}$) yields 98 helpful, 26 harmful, and 496 neutral segments. After long \Sl{} segments ($\ge 6\,\mathrm{s}$), 68 were helpful and 0 harmful; after 2--6\,s segments in \Tf{} and \Ss{}, 0 were helpful and 11 harmful; in \Tb{} ($7.9\,\mathrm{s}$ segments), 18 were helpful and 10 harmful. Offline pre-apply raw VIO error separated all 26 harmful segments (median $1.79\,\mathrm{cm}$, $[0.283, 3.25]\,\mathrm{cm}$) from all 98 helpful segments (median $33.3\,\mathrm{cm}$, $[3.44, 277]\,\mathrm{cm}$) around $\approx 3.3\,\mathrm{cm}$ (\cref{fig:diagnostics}c). Across locally verified EV4 and SWP5 run records, three map-on task non-completions (\texttt{failed=false}, not hard failures) were paired with map-off completion: EV4 \Tb{} scenes 20300102 and 20300118 (exec~7) and SWP5 \Ss{} scene 20300158 (exec~7).

(The EuRoC crossover chart appears directly in \cref{fig:euroc_main} of \cref{sec:costs}.)

\begin{table}[h]
\centering
\caption{EuRoC MAV held-out open-loop estimator replay (72 windows across 7 unseen sequences, 18 per duration; frozen summary \texttt{GATE100\_TEST\_SUMMARY.md}). Penalized 10\,s RMSE medians and lower/higher window counts (within-window thresholded comparison against raw Basalt VIO; remaining windows are neutral or failed).}
\label{tab:euroc}
\footnotesize
\begin{tabular}{lccc}
\toprule
Blackout & Failures (Basalt / no gate / gate) & Median penalized RMSE (m) & Gated vs.\ Basalt (lower / higher) \\
\midrule
1\,s & 0 / 0 / 0 & 0.092 / 0.145 / 0.160 & 3 / 9 (of 18) \\
2\,s & 0 / 0 / 0 & 0.114 / 0.111 / 0.112 & 6 / 3 (of 18) \\
4\,s & 0 / 0 / 0 & 0.196 / 0.162 / 0.173 & 6 / 2 (of 18) \\
8\,s & 10 / 11 / 11 & 0.449 / 0.441 / 0.421 & 3 / 0 (of 7 jointly scored) \\
\bottomrule
\end{tabular}
\end{table}

\textbf{Real-data EuRoC replay.} Seven held-out EuRoC MAV sequences~\citep{burri2016euroc} (\texttt{MH\_03/04/05}, \texttt{V1\_02/03}, \texttt{V2\_02/03}) were replayed with injected 1, 2, 4, and 8\,s stereo blackouts (72 held-out windows, open-loop estimator layer; 48 windows on four other sequences were used earlier for localizer development). Composition uses archived \texttt{oldmap\_compose\_gate.py} (0.05\,m confirm-3, 10\,Hz / 1\,s linear fade, 100-point gate vs.\ 0) scored via \texttt{eur\_test\_eval.py} calling \texttt{degradation.score}; \cref{tab:euroc} reports the frozen summary (\texttt{GATE100\_TEST\_SUMMARY.md} / \texttt{EUROC\_TEST\_RESULT.md}; raw per-window JSONs and \texttt{degradation.py} reside on cold archive): after 1\,s blackouts, three windows improve and nine worsen (median penalized RMSE $0.092 \to 0.160\,\mathrm{m}$); after 4\,s, six improve and two worsen ($0.196 \to 0.173\,\mathrm{m}$); after 8\,s, three improve and zero worsen among seven jointly scored windows (e.g.\ \texttt{MH\_03} $0.381 \to 0.191\,\mathrm{m}$, \texttt{MH\_05} $0.529 \to 0.222\,\mathrm{m}$) while total failures move from 10/18 to 11/18 (\texttt{V2\_03} $0.44\,\mathrm{m}$ vs.\ $1.04\,\mathrm{m}$ gated / $1.20\,\mathrm{m}$ ungated). Real frames observe abundant pre-outage points, so the 100-point gate altered eligibility in at most one window per duration (12/12, 10/10, 9/8, 10/9). This open-loop estimator replay reproduces the simulation bias--drift crossover on real camera-IMU data, motivating drift-conditioned map gating for future deployment.

\section{Compute Footprint}
\label{app:timeline}

(Archived EV4 full-system flight timelines appear directly in \cref{fig:timeline} of \cref{sec:mechanism}.)

\textbf{Compute.} Closed-loop batches ran ten concurrent flights on a workstation-class GPU with a 16-core CPU allocation (4--7\,min wall time per flight under shared load; EV3 ran identical code on an earlier GPU generation). EV4 required $\approx 21.7\,\mathrm{h}$, EV6 $\approx 2.9\,\mathrm{h}$, and the EuRoC held-out replay $\approx 3.3\,\mathrm{h}$ of shared CPU time. Under 10-way concurrent workstation load, logged median execution times were $\approx 18\,\mathrm{ms}$ for MPPI and $\approx 76\,\mathrm{ms}$ per frame for ORB-SLAM3.

\end{document}